\documentclass[journal,twoside,web]{ieeecolor}
\usepackage{generic}
\usepackage{cite}
\usepackage{amsmath,amssymb,amsfonts}
\usepackage{algorithmic}
\usepackage{graphicx}
\usepackage{algorithm,algorithmic}
\usepackage{hyperref}
\hypersetup{hidelinks=true}
\usepackage{textcomp}
\usepackage{booktabs}
\usepackage{multirow}
\begin{document}
\title{CodePhys: Robust Video-based Remote Physiological Measurement through Latent Codebook Querying}
\author{Shuyang Chu, Menghan Xia, Mengyao Yuan, Xin Liu, \IEEEmembership{Senior Member, IEEE}, Tapio Seppänen,  {Guoying~Zhao}, \IEEEmembership{Fellow, IEEE}, and Jingang Shi, \IEEEmembership{Member, IEEE}
\thanks{The manuscript is submitted on Oct. 18, 2024. This work was supported by the National Natural Science Foundation of China (No. 62311530046, 62002283, 62171309), the Research Council of Finland (No. 357137), the Research Council of Finland Academy Professor project EmotionAI (No. 336116, 345122, 359854), and the University of Oulu \& Research Council of Finland Profi 7 (No. 352788).}
\thanks{S. Chu, M. Yuan, and J. Shi are with the
School of Software Engineering, Xi’an Jiaotong University, Xi’an, China. (e-mail:\{xjtucsy,yuanmengyao\}@stu.xjtu.edu.cn, jingang.shi@hotmail.com). M. Xia is with the Tencent AI Lab, Shenzhen, China (e-mail: menghanxyz@gmail.com).}
\thanks{X. Liu is with the Computer Vision and Pattern Recognition Laboratory, Lappeenranta-Lahti University of Technology LUT, 53850 Lappeenranta, Finland (e-mail: xin.liu@lut.fi).}
\thanks{T. Seppänen and G. Zhao are with the Center for Machine Vision and Signal Analysis, University of Oulu, Finland (e-mail: \{tapio.seppanen,guoying.zhao\}@oulu.fi). The first two authors contributes equally. Corresponding author: Jingang Shi}
}
\maketitle

\begin{abstract}
Remote photoplethysmography (rPPG) aims to measure non-contact physiological signals from facial videos, which has shown great potential in many applications. Most existing methods directly extract video-based rPPG features by designing neural networks for heart rate estimation. Although they can achieve acceptable results, the recovery of rPPG signal faces intractable challenges when interference from real-world scenarios takes place on facial video. Specifically, facial videos are inevitably affected by non-physiological factors (e.g., camera device noise, defocus, and motion blur), leading to the distortion of extracted rPPG signals. Recent rPPG extraction methods are easily affected by interference and degradation, resulting in noisy rPPG signals. In this paper, we propose a novel method named CodePhys, which innovatively treats rPPG measurement as a code query task in a noise-free proxy space (i.e., codebook) constructed by ground-truth PPG signals. 
{We consider noisy rPPG features as queries and generate high-fidelity rPPG features by matching them with noise-free PPG features from the codebook. Our approach also incorporates a spatial-aware encoder network with a spatial attention mechanism to highlight physiologically active areas and uses a distillation loss to reduce the influence of non-periodic visual interference.} Experimental results on four benchmark datasets demonstrate that CodePhys outperforms state-of-the-art methods in both intra-dataset and cross-dataset settings. 
\end{abstract}


\section{Introduction}
\label{sec:introduction}

\begin{figure}[t]
\centering
\includegraphics[width=1\columnwidth]{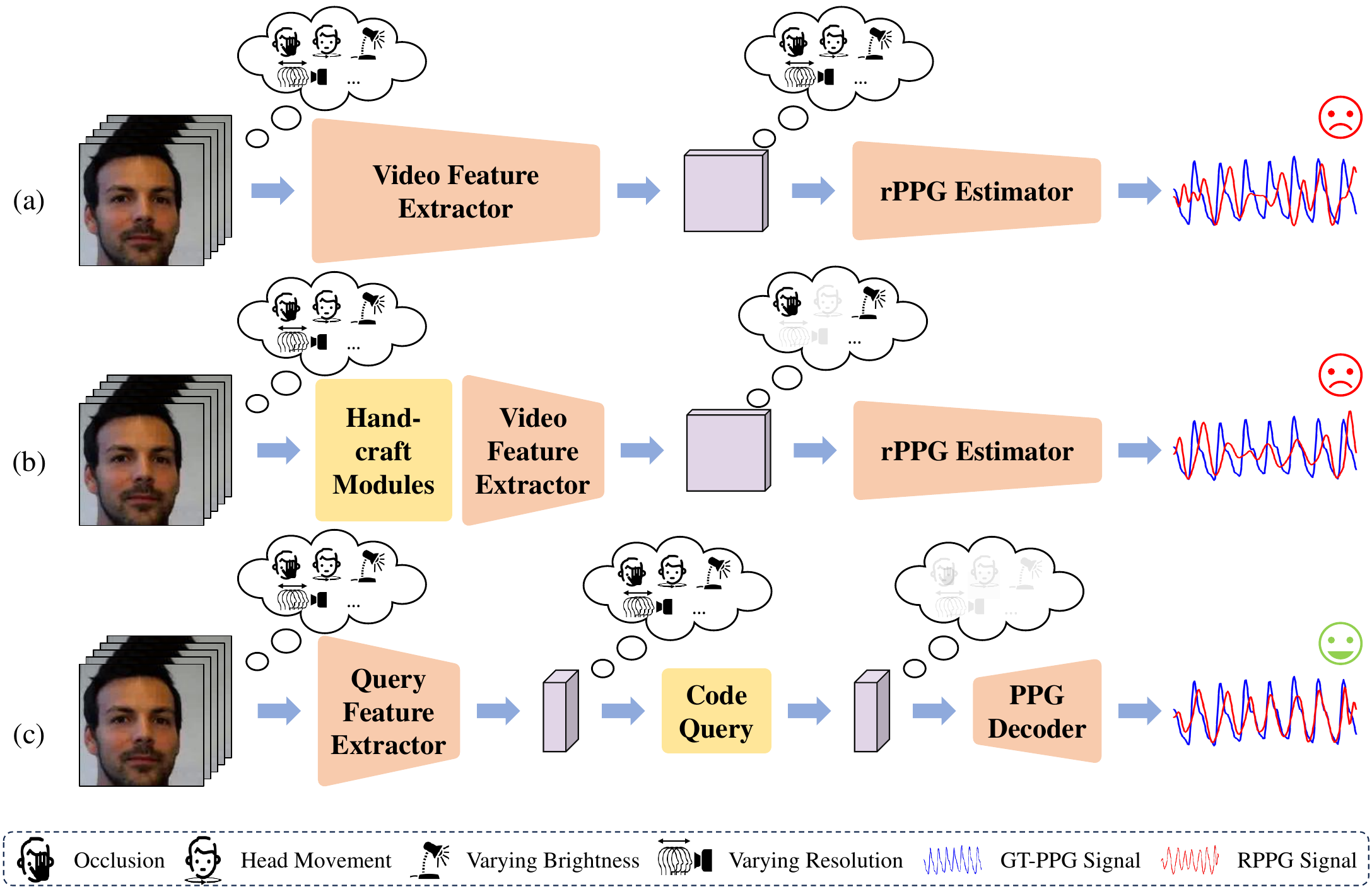} 
\vspace{-1.5em}
\caption{Conceptual comparison of CodePhys and existing typical rPPG measurement methods. (a) Directly extracting rPPG signals from the video. {(b) Designing hand-crafted modules to eliminate specific visual interference, taking the typical method \cite{LiMotionRobust2023} as an example.} (c) Learning a codebook to eliminate all types of visual interference. Here the cloud symbolizes the visual interference present in the video or in features, and the yellow module signifies the capability to remove such interference.}
\label{fig:concept}
\vspace{-1.5em}
\end{figure}

\IEEEPARstart{P}{hysiological} signals are crucial indicators for monitoring human health status. Traditional methods for tracking heart activities and corresponding physiological signals primarily involve the use of electrocardiography (ECG) and photoplethysmography (PPG). However, the complex setup operations and limited scalability of ECG and PPG sensors present practical challenges to real-world applications in daily life. To alleviate these limitations, remote photoplethysmography (rPPG) measurement, which aims to capture heart activities without physical contact, has gained increasing attention in recent years.

Traditionally, rPPG measurement methods often separate rPPG signals from videos using signal processing techniques. They can be divided into two categories: 1) blind signal decomposition techniques such as ICA \cite{PohICA2010} and PCA \cite{LewandowskaPCA2011}, and 2) skin reflection models such as CHROM \cite{DeCHROM2013}, POS \cite{WangPOS2017}, and APOS \cite{ZhangAPOS2024}. Currently, some of these traditional methods have been applied in the field of driving monitoring \cite{GongQSPS2024} and palm-based rPPG measurement \cite{LianPalm2023}. However, these methods work under strong assumptions, limiting their applicability in real-world scenarios with uncertain factors such as varying brightness and head movements. Niu \textit{et al.} propose RhythmNet \cite{NiuRhythm2020} to extract rPPG signals from empirically hand-crafted video features, which firstly constructs the spatio-temporal map (STMap) through feature extraction on different facial regions. Subsequently, non-end-to-end STMap-based methods are developed to alleviate noise by disentangled learning and adversarial learning, such as CVD \cite{NiuCVD2020} and Dual-GAN \cite{LuDualGAN2021}. Recently, with the advancement of deep learning in video understanding, researchers have designed end-to-end networks to directly extract rPPG signals from facial videos using convolutional neural networks (CNNs) \cite{YuPhysNet2019, CINrPPG2023} and Transformers \cite{YuPhysFormer2022, shao2023tranphys}.

The rPPG measurement attempts to mine subtle periodic clues of blood volume from light absorption variation of local skin. However, the magnitude of light absorption variation w.r.t. blood volume is too subtle to be visible, making it susceptible to visual interference such as defocusing, motion blur, varying resolutions, and occlusion. As shown in Fig. \ref{fig:concept}(a), most approaches have overlooked designing suitable solutions to tackle visual interference. As they extract the subtle rPPG signals by naively stacking 3D convolutional layers \cite{YuPhysNet2019,LiuEfficientPhys2021} or Transformer blocks \cite{LuNASHR2021}, these methods may produce degraded rPPG signals due to the above negative factors. As shown in Fig. \ref{fig:concept}(b), some methods attempt to design hand-crafted modules to address specific visual interference such as head movements \cite{LiMotionRobust2023}, varying brightness \cite{CaoSTPhys2024}, and occlusions \cite{WeiConDiff2024}. However, these methods heavily rely on the designed paradigms, and they are susceptible to interference outside of the assumption.

Recently, dictionary learning has been employed to extract a complete set of basis vectors from high-quality image data, thereby creating a dictionary for a sparse representation of the image data. This dictionary can be regarded as a codebook, where each codebook item stores image features with fine details and textures, enabling the restoration of degraded images \cite{JoPSISRULT2021,OordVQVAE2017}. Furthermore, some researchers have applied the dictionary learning to the image generation \cite{EsserVQGAN2021}, image super-resolution \cite{ZhouCodeFormer2022}, and speech-driven 3D facial animation \cite{XingCodetalker2023}. Inspired by these advancements, we propose to decompose the measured noise-free PPG signals to obtain a set of basis vectors that form a codebook, as shown in Fig. \ref{fig:concept}(c). Since the extraction process of rPPG signals is inevitably affected by external visual interference, the noise-free PPG features in the codebook help to alleviate the impact of visual interference on rPPG features. To estimate the codebook, an encoder-decoder structure is adopted to reconstruct ground-truth PPG (GT-PPG) signals. The latent features decomposed by the encoder are combined to acquire the codebook. Since the latent features sampled from the codebook can be reconstructed into high-fidelity PPG signals by the decoder, the obtained codebook can be regarded as a proxy space of PPG signals. Meanwhile, the noise-free latent PPG features serve as prior knowledge to correct the degraded rPPG features when unexpected visual interference accumulates. Based on the noise-free codebook, we propose to transform rPPG measurement into a code query task within the proxy space. In complex real-world scenarios, this mechanism matches the degraded rPPG features with the most similar noise-free PPG feature items in the codebook and corrects the corresponding degraded rPPG features to obtain high-fidelity rPPG signals.

Based on the learned codebook, we propose a code-query-based two-stage model for rPPG measurement to address the signal distortion problem resulting from visual interference, called \textit{CodePhys}, which shows superior robustness to degradation and presents high-fidelity in rPPG recovery. Stage I of CodePhys aims to learn the latent representation of GT-PPG signals, while Stage II aims to recover the rPPG signals by querying the noise-free latent representation with the facial videos. Specifically, in Stage I, we utilize a signal autoencoder framework to reconstruct GT-PPG signals. Subsequently, we construct a set of complete basis vectors from the noise-free latent representation, which form our codebook. The acquired codebook, composed of noise-free PPG features, effectively corrects noisy rPPG features by leveraging the inherent periodic information of PPG signals. In Stage II, we transform the rPPG measurement task into a code query task. Given the input facial video, we customize a spatial-aware encoder for mapping the video to corresponding rPPG features. The spatial-aware encoder explicitly captures spatial information, generating a spatial attention map to identify the importance of physiological signals from different facial regions. To mitigate the impact of visual interference on rPPG features, we implement a PPG feature extractor to distill features from GT-PPG signals, thereby enhancing the periodicity of rPPG features. These refined features act as query features, which we use to identify the closest noise-free PPG feature in the codebook. This process corrects the unknown degradation in the query features. Finally, the rPPG signal is recovered from the corrected noise-free query features by the pre-trained signal decoder in Stage I.


We conduct experimental evaluation on four benchmark datasets, in both intra-dataset and cross-dataset settings. It shows that CodePhys significantly outperforms state-of-the-art methods across multiple metrics. CodePhys achieves strong performance even in challenging cases of diverse video quality degradation, which indicates its potential for a wide range of applications. In summary, the contributions of our work are as follows:

\begin{itemize}
    \item  We propose CodePhys, a novel framework based on codebook querying, designed to mitigate visual interference in rPPG measurement by modeling an equivalent noise-free PPG latent representation. This innovative approach allows CodePhys to provide more accurate and robust rPPG estimation, even in the presence of visual artifacts that typically affect existing methods.
    \item We propose a soft feature distillation loss function that enhances the robustness of video features by distilling the periodicity information from high-quality GT-PPG signals. This process effectively condenses the essence of the GT-PPG signals into the feature extraction pipeline, ensuring that the learned rPPG features are periodically consistent and resistant to non-physiological interference. 
    \item The proposed approach serves as a robust, plug-and-play framework that can be integrated into any end-to-end rPPG measurement networks with minimal modifications. This flexibility allows existing methods to leverage the benefits of advanced noise reduction capabilities without extensive overhauls.
    \item The extensive experiments demonstrate that CodePhys achieves the best performance among state-of-the-art methods on various datasets even in scenarios with serious degradation.
 \end{itemize}

\section{Related Work}
\subsection{Remote Physiological Measurement}
\label{subsec:physiological_measurement}

Since the successful extraction of rPPG signals from facial videos by \cite{VerkruysseGreen2008}, plenty of rPPG measurement approaches have been developed. Initially, traditional signal processing methods based on signal decomposition \cite{PohICA2010} were developed to extract rPPG signals from facial videos, recovering fundamental physiological signals through Blind Source Separation (BSS) techniques such as Independent Component Analysis (ICA) \cite{McDuffANMPM2011,PohICA2010} and Principal Component Analysis (PCA) \cite{LewandowskaPCA2011}. However, BSS-based techniques did not take into account the physical and optical properties of the skin. To leverage prior knowledge of the physiological waveform dynamics, several methods were proposed based on signal projection or color space decomposition. For instance, CHROM \cite{DeCHROM2013} eliminated specular reflections in videos by transforming the face video into a linear combination of chrominance-signals. POS \cite{WangPOS2017} proposed a physically grounded demixing approach by defining a plane orthogonal to the skin color space. To further enhance the generalization of POS, Zhang \textit{et al.} proposed APOS \cite{ZhangAPOS2024}, which can adaptively determine the signal plane used in POS, thereby improving the robustness against complex scenarios. However, all these signal processing methods struggled to effectively separate noise from different sources and neglected a significant amount of spatio-temporal and chromatic space information \cite{McDuffRPPGSURVEY2023}. 

In recent years, deep learning (DL) models have gained prominence in rPPG measurement due to their exceptional nonlinear fitting capabilities and spatio-temporal representation capacities. Among these methods, early end-to-end spatio-temporal networks captured physiological signals in face videos by designing and combining 3D convolution modules \cite{liu2020Deeprppg, CINrPPG2023}. CIN-rPPG \cite{CINrPPG2023} proposed utilizing both the channel and spatial interactions to extract video features for accurate rPPG measurement. As the development of the Transformer in computer vision field, some researchers transferred Vision Transformer (ViT) \cite{AlexeyViT2021} or its variants \cite{LiuSwin2021} to the rPPG measurement task for capturing global spatio-temporal context information. Yu \textit{et al.} first proposed PhysFormer \cite{YuPhysFormer2022} and PhysFormer++ \cite{YuPhysformer++2023} to leverage the long-range sequence modeling ability of Transformer architecture for rPPG measurement. Afterward, Dual-TL \cite{Dual-TL2024} used a hybrid spatio-temporal Transformer-based architecture to enhance the modeling capability for long video sequence dependencies. Some other approaches \cite{shao2023tranphys, LiurPPGMAE2023} were inspired by the Masked Autoencoder (MAE) \cite{HeMAE2022} paradigm, which enhanced the quality of the estimated rPPG signals by pre-training Transformer models on large datasets (e.g., VIPL-HR \cite{NiuVIPL2018}) as prior knowledge. Additionally, some researchers proposed to manually craft spatio-temporal map (STMap) to inherit the self-similar prior of facial videos by averaging pixels in different facial regions of interest (ROIs) \cite{NiuRhythm2020,NiuCVD2020}, but such self-similarity was hard to obtain when serious head movement happened \cite{LiurPPGMAE2023}. In general, these methods ignored the effect of visual interference in the real-world, which may produce degraded rPPG signals and harm the fidelity of heart rate measurement.

To alleviate visual interference induced by head movement, Li et al. \cite{LiMotionRobust2023} manually designed Physiological Feature Extraction (PFE) and Temporal Face Alignment (TFA) modules for capturing facial motion and resolution variations. STPhys \cite{CaoSTPhys2024} built a CNN-based spatio-temporal model to enhance video quality and achieve high accurate rPPG measurement in low-light conditions. ConDiff-rPPG \cite{WeiConDiff2024} used the diffusion model to restore the damaged STMap, thereby improving the robustness under heavy head movements and occlusions. Lastly, MotionMatters \cite{MotionMatters2024} proposed to augment the facial videos with existing motion magnification approaches, thereby enhancing the model's robustness against head motion after training. However, these models heavily relied on these modules tailored for the specific type of interference. As a result, they performed poorly when faced with other unexpected visual disturbances. To address these limitations, we propose to decompose the noise-free GT-PPG signals into a latent feature representation. These noise-free PPG features are combined into a codebook that can correct degraded rPPG features, regardless of the type of visual interference that accumulates. With the learned noise-free codebook, we transform rPPG measurement into a code query task to mitigate the visual interference of non-physiological information contained in the video.

\subsection{Discrete Representation Learning}
\label{subsec:prior_learning}

Discrete representation with learned dictionary has demonstrated its superiority in image restoration tasks, such as super-resolution \cite{YangISRVSR2010, TimofteANRFESR2013, AhmedCSCUW2019, GuoTCSVT24} and denoising \cite{EladIDVSROLD2006}, since the fine details and textures could be well-preserved in the dictionary. The concept of leveraging pre-learned prior knowledge to restore degraded images and videos further drove advancements in deep learning-based image restoration \cite{JoPSISRULT2021} and synthesis methods \cite{OordVQVAE2017, RazaviVQVAE22019, EsserVQGAN2021}. VQ-VAE \cite{OordVQVAE2017} was the first to use a highly compressed representation (i.e., codebook) learned by the vector quantized autoencoder model for image synthesis. VQ-GAN \cite{EsserVQGAN2021} further improved the quality of synthesized images while significantly reducing the codebook size by using adversarial loss and perceptual loss. Recently, codebook-based discrete prior representation learning has been exploited in other fields. UniColor \cite{HuangUniColor2022} achieved robust image colorization by disentangling and quantizing chroma representation from a continuous grayscale image with a codebook. CodeFormer \cite{ZhouCodeFormer2022} decomposed high-resolution face images into a codebook with VQ-VAE, then achieved image super-resolution by replacing low-resolution face image features with the decomposed codebook items. Similarly, GSS \cite{ChenGSS2023} constructed a codebook consisting of real semantic masks with a VQ-VAE and queried the codebook items with an additional image encoder to assist semantic segmentation. 

Due to the ability of the discrete prior representation to preserve pure knowledge details, more recent works have attempted to extend discrete representation learning to modalities outside of images. Evonne et al. \cite{NgL2L2022} proposed to disentangle facial actions (expression coefficients and 3D head rotations) into an action unit codebook, which generated realistic facial actions of the listener w.r.t. the facial actions and audio of the speaker. PCT \cite{GengPCT2023} posited that the coordinate vectors of body joints can be equivalently represented by multiple discrete tokens, while these tokens together formed a codebook prior to human poses. This formulation allowed for performing robust human pose estimation by encoding image features to query the learned codebook items. Inspired by the 3D Face Morphable Model (3DMM) \cite{Li3DMM2017}, CodeTalker \cite{XingCodetalker2023} represented general facial expressions with a finite discrete codebook as a prior for facial motions, thereby proposing a temporal autoregressive model for speech-conditioned facial motion synthesis. Taking the recent advancement of discrete representation learning, we first explore to design a adaptive noise-free codebook for PPG features in order to achieve robust rPPG measurement in this paper.

\begin{figure*}[t]
\centering
\includegraphics[width=1\linewidth]{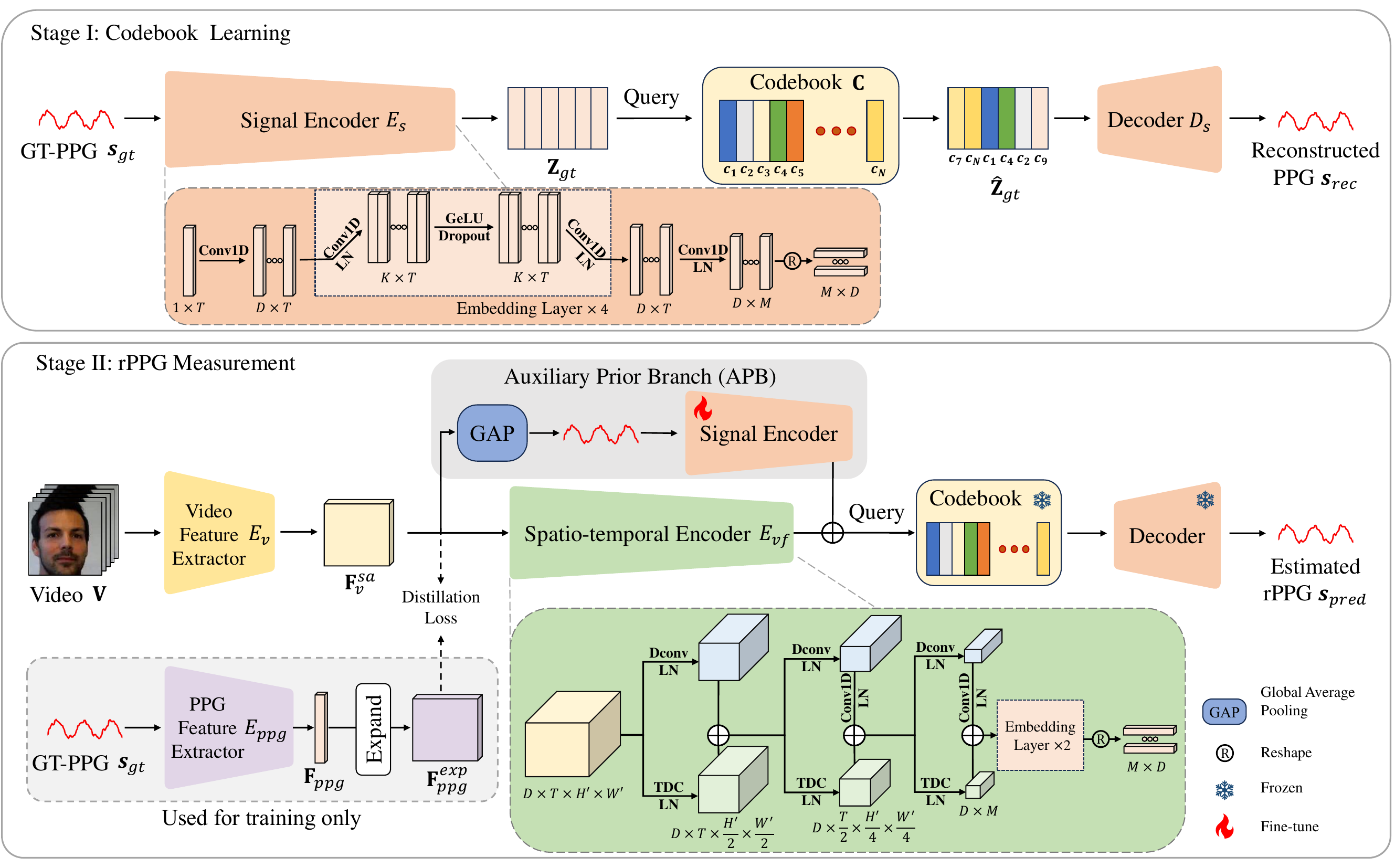 }
\caption{Overview of CodePhys. In Stage I, a codebook consisting of noise-free PPG features is learned by reconstructing GT-PPG signals, which is treated as the prior. In Stage II, the spatial-aware encoder (composed of a video feature extractor and a spatio-temporal encoder with an auxiliary prior branch) queries corresponding PPG features with respect to the input video. Subsequently, the decoder network reconstructs the estimated rPPG signal from the queried features. The embedding layer in Stage II is the same as that in Stage I. The DConv, TDC, Conv1D, and LN represent deformable convolution \cite{DaiDeformCNN2017}, temporal difference convolution \cite{YuAuto2020}, 1D convolution, and layer normalization, respectively.}
\label{fig:overview}
\end{figure*}
\section{Methodology}
\label{sec:method}

We propose CodePhys, a codebook-based framework for rPPG measurement from facial videos. The learning of CodePhys is divided into two stages, as shown in Fig. \ref{fig:overview}. In the codebook learning stage (Stage I), we employ a signal autoencoder to reconstruct GT-PPG signals, thereby constructing a noise-free codebook. In the rPPG measurement stage (Stage II), we specifically design a spatial-aware encoder network to extract rPPG features from the input facial video, and the matched noise-free PPG feature items in the learned codebook are then used for rPPG reconstruction. The detailed model components and training loss functions are described below. 

\subsection{Codebook Learning (Stage I)}
\label{subsec:stage1}

Complex real-world scenarios make it challenging to extract high-fidelity PPG signals from facial videos. To overcome the difficulty, we aim to construct an equivalent noise-free latent representation of PPG signals. We achieve this by employing a signal autoencoder network that decomposes GT-PPG signals into latent PPG features. These noise-free features are then combined to form a codebook, as shown in the codebook learning stage of Fig. \ref{fig:overview}.

\subsubsection{Signal autoencoder network}
\label{subsec:signal_vqvae}
Given a GT-PPG signal $\boldsymbol{s}_{gt}\in \mathbb{R}^T$ (where $T$ is the signal length), we use a signal encoder $E_s$ to transform $\boldsymbol{s}_{gt}$ into $M$ temporally compressed PPG features, each of which has $D$ dimensions:
\begin{equation}
\begin{split}
  \label{eq:obtainZ}
  \mathbf{Z}_{gt}=(\boldsymbol{z}_1, \boldsymbol{z}_2, \dots, \boldsymbol{z}_M)^\mathrm{T}=E_s(\boldsymbol{s}_{gt}).
\end{split}
\end{equation}

Fig. \ref{fig:overview} shows the network structure of the signal encoder $E_s$. Initially, the PPG signal is processed by a 1D convolution with a kernel size of 1$\times$1 to augment feature representation. Subsequently, the features are funneled through a sequence of embedding layers, which effectively integrate the features across various temporal positions. Finally, we extract $M$ PPG features through a large kernel 1D convolution (kernel size 1$\times$5 with stride 4). Note that the parameter ${M}$ is one-fourth of $T$ and the parameter $K$ is twice $D$, enabling each PPG feature $\boldsymbol{z}_i\in\mathbb{R}^{D}$ to approximately encode the temporal context within a certain time range.

To explicitly store the noise-free latent representation of PPG signals, we define a latent embedding space by a codebook $\mathbf{C}=(\boldsymbol{c}_1, \boldsymbol{c}_2, \dots, \boldsymbol{c}_N)^\mathrm{T}\in \mathbb{R}^{N \times D}$ with $N$ items, where each item $\boldsymbol{c}_i \in \mathbb{R}^{D}$ is initialized randomly and optimized through GLO strategy \cite{bojanowski2017GLO}. 

The visual interference in $\mathbf{Z}_{gt}$ could be mitigated by replacing  $\mathbf{Z}_{gt}$ with the noise-free PPG features in the codebook, which is the key insight of CodePhys. The code query process could be divided into two steps: (i) obtaining the coordinates of the most similar codebook items in $\mathbf{C}$ corresponding to each item in $\mathbf{Z}_{gt}$, and (ii) replacing $\mathbf{Z}_{gt}$ with the obtained coordinates to get the quantized PPG representation $\hat{\mathbf{Z}}_{gt}$.

First, we define a query function $\mathbf{Q}_{gt}=\text{QUERY}(\mathbf{Z}_{gt}, \mathbf{C})\in \mathbb{R}^{M \times N}$, which maps each token in $\mathbf{Z}_{gt}$ to the coordinate of its nearest noise-free PPG feature in codebook $\mathbf{C}$. The query coordinates for all features could be described as $\mathbf{Q}_{gt}=(\boldsymbol{q}_1, \boldsymbol{q}_2, \dots, \boldsymbol{q}_M)^{\mathrm{T}}$, where $\boldsymbol{q}_i\in \{0,1\}^{N}$ is a one-hot vector obtained by $\boldsymbol{z}_i$ with the following formulation:
\begin{equation}
\begin{split}
  \label{eq:obtainQ}
  \boldsymbol{q}_{ij} = \begin{cases}
   \ 1& \mathrm{if}\ \  j = \arg\min\limits_{k}\left\|{\boldsymbol{z}_i - \boldsymbol{c}_k}\right\|_2\\
   \ 0& \mathrm{else},
  \end{cases}
\end{split}
\end{equation}
where $\boldsymbol{q}_{ij}$ indicates the $j$-th value in $\boldsymbol{q}_i$.

Second, we can simply multiply the query coordinates $\mathbf{Q}_{gt}$ with the codebook $\mathbf{C}$ to describe the code query process as $\hat{\mathbf{Z}}_{gt}=\mathbf{Q}_{gt}\cdot\mathbf{C}\in \mathbb{R}^{M \times D}$. The signal decoder $D_s$ then reconstructs the GT-PPG signal $\boldsymbol{s}_{rec}\in\mathbb{R}^{T}$ from the quantized PPG representation $\hat{\mathbf{Z}}_{gt}$:
\begin{equation}
    \label{s-rec}
    \boldsymbol{s}_{rec} = D_s(\hat{\mathbf{Z}}_{gt}),
\end{equation}
where the signal decoder $D_s$ mirrors the structure of $E_s$ in reverse to ensure its ability to reconstruct the PPG signal.

\subsubsection{Codebook settings}
\label{subsec:codebook_setting}

The capacity of the codebook is crucial for the accurate representation of the PPG data space. Specifically, insufficient capacity lacks the necessary flexibility to accurately capture the periodicity of PPG signals, while overly abundant capacity eases the reconstruction but challenges the code prediction due to the ambiguity of redundant PPG feature items. Empirically, we use a codebook with $N=64$ PPG feature items (codes), and each PPG feature item has a dimension of $D=64$ in all the experiments.

\subsubsection{Training objectives}
\label{subsec:codebook_train}
We supervise the reconstruction of GT-PPG signals in two aspects. We first adopt two temporal domain losses (i.e., Mean Square Error (MSE) loss and negative Pearson (NP) loss \cite{YuPhysNet2019}) to supervise the reconstruction quality:
\begin{equation}
\begin{split}
\label{eq:l_rec}
  \mathcal{L}_{rec} &= \mathrm{MSE}(\boldsymbol{s}_{gt},\boldsymbol{s}_{rec}) + \mathrm{NP}(\boldsymbol{s}_{gt},\boldsymbol{s}_{rec}).\\
\end{split}
\end{equation}

We further adopt one code-level loss to minimize the quantization loss (i.e., the distance between $\mathbf{Z}_{gt}$ and the codebook items):
\begin{equation}
\begin{split}
\label{eq:l_feat}
   \mathcal{L}_{feat} &= \left\|{\mathrm{sg}(\mathbf{Z}_{gt}) - \hat{\mathbf{Z}}}_{gt}\right\|_2^2 + \delta\left\|{\mathbf{Z}_{gt} - \mathrm{sg}(\hat{\mathbf{Z}}_{gt})}\right\|_2^2,
\end{split}
\end{equation}
where $\mathrm{sg}(\cdot)$ denotes the stop-gradient operator and $\delta=0.25$ is a weight coefficient to modulate the update rates of the encoder and codebook learning. 

Note that the first item in $\mathcal{L}_{feat}$ reduces quantization loss by optimizing the codebook $\mathbf{C}$ without updating the encoder $E_s$, while the second item does the opposite. Considering that the input signal is constantly changing, this design prevents the output of $E_s$ from frequently fluctuating between different PPG feature items in $\mathbf{C}$ \cite{OordVQVAE2017}.

As the $\mathrm{argmin}$ operation in the code query process Eq. (\ref{eq:obtainQ}) is not differentiable, we use the straight-through gradient estimator \cite{EsserVQGAN2021} to solve the problem. We jointly optimize the encoder $E_s$, decoder $D_s$, and codebook $\mathbf{C}$ by the final objective of Stage I:
\begin{equation}
\begin{split}
  \label{eq:l_code}
  \mathcal{L}_{code} = \mathcal{L}_{rec}+\mathcal{L}_{feat}.
\end{split}
\end{equation}

Note that we do not use extra data when training Stage I to avoid data leakage. Only the GT-PPG signals from the training set are used in this stage.

\subsection{RPPG Measurement (Stage II)}
\label{subsec:stage2}
As the signal autoencoder network has been pre-trained with noise-free GT-PPG signals in Stage I, we posit that the learned codebook $\mathbf{C}$ and decoder $D_s$ collectively encapsulate the noise-free prior knowledge of PPG signals. To leverage this prior knowledge, we freeze $\mathbf{C}$ and $D_s$, transforming the rPPG measurement into a query code prediction task. Therefore, the primary procedures of Stage II involve extracting rPPG features from the input video and querying the closest noise-free PPG features to them within the codebook. These queried features could be used to reconstruct the rPPG signal via the pre-trained signal decoder $D_s$.

It is challenging to extract the rPPG features from the facial video due to various visual information beyond the subtle physiological features. To solve this challenge, we propose a novel spatial-aware encoder network that comprises two components: a video feature extractor $E_v$ with a Spatial Attention Mechanism (SAM), and a spatio-temporal encoder $E_{vf}$ with an Auxiliary Prior Branch (APB). We adopt $E_v$ to model spatial information and extract the video feature, and $E_{vf}$ to map the video feature to rPPG features. In addition, we further employ a PPG feature extractor $E_{ppg}$ to offer feature level supervision in the way of knowledge distillation. The feature level supervision improves the alignment of periodicity between the video feature and the GT-PPG signals.

\subsubsection{Spatial-aware encoder}
\label{subsec:spatial-aware_encoder}

\begin{figure}[t]
\centering
\includegraphics[width=1\columnwidth]{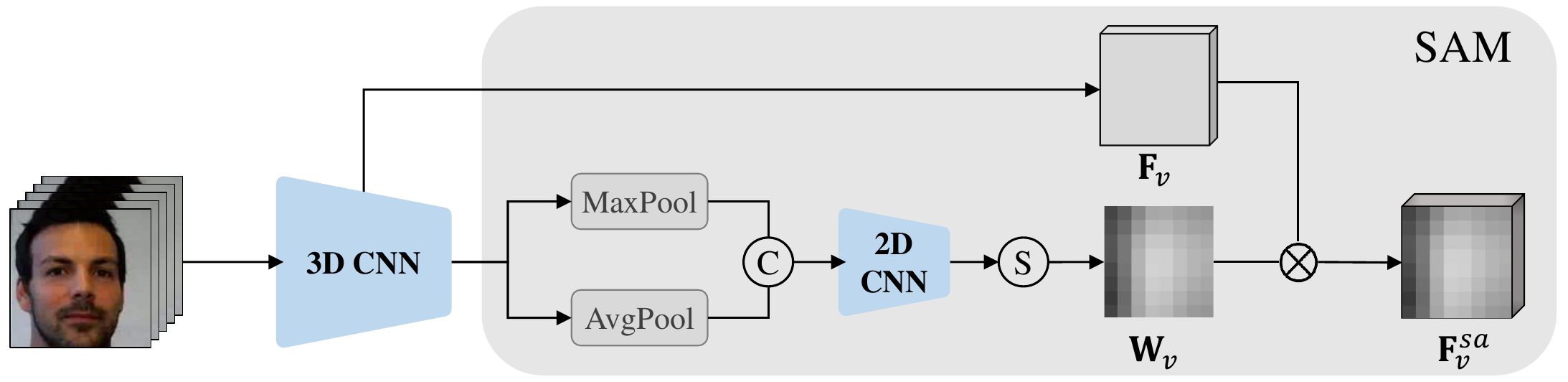}\vspace{-0.5em} 
\caption{Architecture of the video feature extractor $E_v$ with a spatial attention mechanism (SAM) attached. The `S' and `C' denote the sigmoid function and concatenation operation, respectively.}
\label{fig:spatial_attention}
\end{figure}

Considering the imbalance of physiological information depicted in spatial positions of the video, we introduce a spatial attention mechanism in video feature extraction, as illustrated in Fig. \ref{fig:spatial_attention}. Given an input video clip $\mathbf{V}\in\mathbb{R}^{3\times T\times H\times W}$ and the ground-truth PPG signal $\boldsymbol{s}_{gt}\in\mathbb{R}^{T}$, we first extract the initial video feature $\mathbf{F}_{v}\in\mathbb{R}^{D\times T\times H'\times W'}$ with a 3D CNN block, where $D$, $T$, $H$, $W$ refer to channel, sequence length, height and width, while $H'$ and $W'$ are $H/16$ and $W/16$, respectively. The 3D CNN block starts with an initial 1$\times$5$\times$5 convolution for channel extraction, followed by four stacked 3$\times$3$\times$3 convolutions interleaved with 1$\times$2$\times$2 max pooling layers. Subsequently, the corresponding attention map $\mathbf{W}_v\in\mathbb{R}^{ H'\times W'}$  is obtained through the spatial attention mechanism:
\begin{equation}
    \label{eq:sam}
    \mathbf{W}_v=\sigma(f_{3\times 3}^\text{Conv2D}(\text{MaxPool}(\mathbf{F}_v)\oplus \text{AvgPool}(\mathbf{F}_v))),
\end{equation}
where $\sigma(\cdot)$ and $\oplus$ indicate the $\text{Sigmoid}(\cdot)$ function and concatenation operation, respectively. $\text{MaxPool}(\cdot)$ and $ \text{AvgPool}(\cdot)$ perform pooling $\mathbf{F}_v$ on the temporal domain. $f_{3\times 3}^\text{Conv2D}$ denotes a 2D convolution with the kernel size of 3$\times$3.

We apply the attention map to the initial video feature by a dot product to obtain the spatial-aware video feature $\mathbf{F}_{v}^{sa}\in\mathbb{R}^{D\times T\times H'\times W'}$. The process can be described as $\mathbf{F}_{v}^{sa}=E_v(\mathbf{V})=\mathbf{F}_{v}\otimes\mathbf{W}_v$, where $\otimes$ denotes the Hadamard multiplication.



With the spatial-aware video feature $\mathbf{F}_{v}^{sa}$, we apply a spatio-temporal encoder $E_{vf}$ to encode $\mathbf{F}_{v}^{sa}$ into compact, time-reduced rPPG features $\mathbf{Z}_{rppg\_enc}\in\mathbb{R}^{M\times D}$.  As shown in Fig. \ref{fig:overview}, we incorporate deformable convolution \cite{DaiDeformCNN2017} to track facial movements in the spatial domain and temporal difference convolution \cite{YuAuto2020} to enhance temporal frequency information. The encoding procedures of $E_{vf}$ can be described as:
\begin{equation}
    \begin{split}
        \label{eq:get-rppg-enc}
      &  \mathbf{Z}_1 = \text{LN}(f_{3\times3\times3}^\text{DConv}(\mathbf{F}_{v}^{sa}))+\text{LN}(f_{3\times3\times3}^\text{TDC}(\mathbf{F}_{v}^{sa})), \\
      &  \mathbf{Z}_2 = \text{LN}(f_{1\times1}^\text{Conv1D}(\text{LN}(f_{3\times3\times3}^\text{DConv}(\mathbf{Z}_1))))+\text{LN}(f_{3\times3\times3}^\text{TDC}(\mathbf{Z}_1)), \\
       &  \mathbf{Z}_3 = \text{LN}(f_{1\times1}^\text{Conv1D}(\text{LN}(f_{3\times3\times3}^\text{DConv}(\mathbf{Z}_2))))+\text{LN}(f_{3\times3\times3}^\text{TDC}(\mathbf{Z}_2)), \\
        & \mathbf{Z}_{rppg\_enc} = \text{Reshape}(\text{EL}(\text{EL}(\mathbf{Z}_3)),
    \end{split}
\end{equation}
where the $f_{3\times3\times3}^\text{DConv}$, $f_{3\times3\times3}^\text{TDC}$, and $f_{1\times1}^\text{Conv1D}$ denote the 3D deformable convolution, 3D temporal difference convolution, and 1D convolution with the corresponding kernel size. $\text{Reshape}(\cdot)$ and $\text{LN}(\cdot)$ denote the reshape and layer normalization operation, respectively. $\text{EL}(\cdot)$ is the embedding layer same as that in the signal encoder $E_s$.

Considering that the pre-trained signal encoder $E_s$ can obtain rPPG features with high distribution consistency as the codebook $\mathbf{C}$, we believe that reusing $E_s$ reasonably helps the spatio-temporal encoder $E_{vf}$ overcome the quantization loss more quickly. Therefore, we propose to obtain  a pseudo-periodic signal from the spatial-aware video feature $\mathbf{F}_{v}^{sa}$ through global average pooling (GAP). Then we adopt a signal encoder identical to $E_s$ to assist the spatio-temporal encoder as an auxiliary prior branch (APB) and refer the output of APB as $\mathbf{Z}_{rppg\_apb}\in\mathbb{R}^{M\times D}$, as shown in Fig. \ref{fig:overview}. It is worth noting that the signal encoder in the auxiliary prior branch is pre-trained in Stage I and fine-tuned in Stage II. Thus, the extracted rPPG feature $\mathbf{Z}_{rppg\_apb}$ from the signal encoder has similar distribution with the PPG feature items in the pre-trained codebook $\mathbf{C}$. Since the auxiliary prior branch and the spatio-temporal encoder extract rPPG features from the spatial-aware video feature $\mathbf{F}_v^{sa}$ in parallel, we apply AdaIN to adjust the mean and variance of $\mathbf{Z}_{rppg\_enc}$ in order to keep consistent with $\mathbf{Z}_{rppg\_apb}$:
\begin{equation}
\begin{split}
  \label{eq:z_venc}
  \tilde{\mathbf{Z}}_{rppg\_enc} = &\ \sigma(\mathbf{Z}_{rppg\_apb})\times\frac{\mathbf{Z}_{rppg\_enc}-\mu(\mathbf{Z}_{rppg\_enc})}{\sigma(\mathbf{Z}_{rppg\_enc})} \\ &+\mu(\mathbf{Z}_{rppg\_apb}),
\end{split}
\end{equation}
where $\sigma(\cdot)$ and $\ \mu(\cdot)$ represent the standard deviation and mean of the input, respectively. The final rPPG features are obtained by:
\begin{equation}
    \label{eq:conbine-z-rppg}
    \mathbf{Z}_{rppg} =\tilde{\mathbf{Z}}_{rppg\_enc} + \mathbf{Z}_{rppg\_apb}.
\end{equation}

Given the rPPG features $\mathbf{Z}_{rppg}$ extracted from the input video $\mathbf{V}$, we employ the pre-trained codebook to eliminate the interference and refine the rPPG features. We search the codebook for the noise-free PPG feature items that most closely match the query features and replace them accordingly. With the fixed codebook $\mathbf{C}$ and pre-trained signal decoder $D_s$, we obtain the query coordinates $\mathbf{Q}_{rppg}$ of rPPG features $\mathbf{Z}_{rppg}$ by $\mathbf{Q}_{rppg}=\text{QUERY}(\mathbf{Z}_{rppg}, \mathbf{C})$. The corresponding quantized rPPG features $\hat{\mathbf{Z}}_{rppg}$ could be obtained through $\hat{\mathbf{Z}}_{rppg}=\mathbf{Q}_{rppg}\cdot \mathbf{C}$. Subsequently, the high-fidelity rPPG signal $\boldsymbol{s}_{pred}\in\mathbb{R}^T$ is decoded from the quantized rPPG features:
\begin{equation}
    \label{eq:s-pred}
    \boldsymbol{s}_{pred}=D_s(\hat{\mathbf{Z}}_{rppg}).
\end{equation}


\subsubsection{Soft feature distillation (SFD)}
\label{subsec:sfd}
In rPPG measurement, physiological signals extracted from different facial regions are expected to share the same periodicity. However, it is inevitable to involve various visual interference during video capturing, such as camera device noise, defocus, motion blur, varying illumination, etc. To resist the influence of such non-physiological visual factors, we introduce soft feature distillation (SFD) in the proposed method. SFD employs GT-PPG features to guide the extraction of spatial-aware features from facial videos. By leveraging the periodic characteristics of GT-PPG signals, SFD enhances the robustness of the rPPG measurement process, ensuring that the extracted features are less susceptible to visual artifacts. 

We employ a PPG feature extractor $E_{ppg}$ to extract the GT-PPG feature ${\mathbf{F}}_{ppg} \in \mathbb{R}^{D\times T }$. The network structure of $E_{ppg}$ is essentially identical to the signal encoder $E_s$, with only the last convolutional layer missing. The GT-PPG feature is expanded to yield the feature distillation target  ${\mathbf{F}}_{ppg}^{exp} \in \mathbb{R}^{D\times T \times H' \times W'}$. We design a soft feature distillation loss $\mathcal{L}_{distill}$ to weight the deviation between ${\mathbf{F}}_{ppg}^{exp}$ and $\mathbf{F}_v^{sa}$ in spatial dimension:
\begin{equation}
\label{eq:l_distall}
\begin{split}
  \mathcal{L}_{distill} &= \sum_{h=1}^{H'}\sum_{w=1}^{W'} D(\mathbf{F}_v^{sa}{(h,w)},\mathrm{sg}(\mathbf{F}_{ppg}^{exp}{(h,w)})),
\end{split}
\end{equation}
where $D(\cdot,\cdot)$ indicates the distance function and $\mathrm{sg}(\cdot)$ indicates the stop-gradient operator.  We employ the smooth L1 loss as the distance function. The gradient-stopping operators in $\mathcal{L}_{distill}$ can prevent $\mathbf{F}_{ppg}^{exp}$  from being optimized based on video feature $\mathbf{F}_v^{sa}$.

Note that PPG feature extractor $E_{ppg}$ is just used for training supervision but not used in the inference phase.

\subsubsection{Training objectives} 
\label{codephys_train}
We supervise the estimated rPPG signal in three aspects. First, to optimize the quality of $\boldsymbol{s}_{pred}$, we supervise the signal using both temporal and frequency domain losses:
\begin{equation}
\begin{split}
  \label{eq:l_phy}
  \mathcal{L}_{phy} = \underbrace{\lambda\cdot\mathrm{NP}(\boldsymbol{s}_{pred}, \boldsymbol{s}_{gt})}_{temporal} + \underbrace{\mathrm{CE}(\mathrm{PSD}(\boldsymbol{s}_{pred}),\mathrm{PSD}(\boldsymbol{s}_{gt}))}_{frequency},
\end{split}
\end{equation}
where $\boldsymbol{s}_{pred}$ is obtained by Eq. (\ref{eq:s-pred}), $\lambda=0.1$ is the weight trade-off between temporal and frequency domain, $\mathrm{NP}(\cdot,\cdot)$ is the negative Pearson correlation mentioned before, $\mathrm{PSD}(\cdot)$ denotes the calculation of power spectral density, and $\mathrm{CE}(\cdot,\cdot)$ is the cross-entropy loss.

Second, to optimize the accuracy of code query process, we supervise the prediction of codebook items using two code-level losses: (i) cross-entropy loss to minimize the difference between query coordinates for $\mathbf{V}$ and corresponding label $\boldsymbol{s}_{gt}$, and (ii) L2 loss to minimize the distance between query features $\mathbf{Z}_{rppg}$ and the quantized rPPG features $\hat{\mathbf{Z}}_{rppg}$ from codebook $\mathbf{C}$:
\begin{equation}
\begin{split}
  \label{eq:l_code'}
  \mathcal{L}_{code'} &= \mathrm{CE}(\mathbf{Q}_{rppg}, \mathbf{Q}_{gt}) + \left\|{\mathbf{Z}_{rppg}-\mathrm{sg}(\hat{\mathbf{Z}}_{rppg}})\right\|_2^2,
\end{split}
\end{equation}
where $\mathbf{Q}_{gt}=\text{QUERY}(E_s(\boldsymbol{s}_{gt}),\mathbf{C})$ is the ground-truth query coordinates of $\boldsymbol{s}_{gt}$ in the codebook items. 

Thirdly, to supervise $E_{ppg}$, we process $\mathbf{F}_{ppg}^{exp}$ with the same procedures as the spatial-aware video feature $\mathbf{F}_v^{sa}$ (i.e., Eq. (\ref{eq:get-rppg-enc}), Eq. (\ref{eq:z_venc}), and Eq. (\ref{eq:conbine-z-rppg})). With the obtained PPG features $\mathbf{Z}_{ppg}$ , PPG query coordinates $\mathbf{Q}_{ppg}=\text{QUERY}(\mathbf{Z}_{ppg},\mathbf{C})$, and corresponding quantized PPG features $\hat{\mathbf{Z}}_{ppg}=\mathbf{Q}_{ppg}\cdot \mathbf{C}$, we employ similar strategy as Eq. (\ref{eq:l_code'}) for supervision:
\begin{equation}
\label{eq:l_code^rppg}
\begin{split}
  \mathcal{L}_{code'}^{ppg} = \mathrm{CE}(\mathbf{Q}_{ppg}, \mathbf{Q}_{gt}) + \left\|{\mathbf{Z}_{ppg}-\mathrm{sg}(\hat{\mathbf{Z}}_{ppg}})\right\|_2^2.
\end{split}
\end{equation}

The overall loss for Stage II is governed by multiple loss components that supervise different aspects of the model. To optimize the estimated rPPG signal, $\mathcal{L}_{phy}$ supervises the spatio-temporal encoder (composed of $E_v$ and $E_{vf}$) at the signal level. To optimize the code query process, $\mathcal{L}_{code'}$ and $\mathcal{L}_{code'}^{ppg}$ supervise spatio-temporal encoder and PPG feature extractor $E_{ppg}$ at the code level. Additionally, $\mathcal{L}_{distill}$ supervises the video feature extractor $E_v$ at the feature level. The formulation of the overall loss for Stage II can be expressed as a combination of these individual loss components, where each term is carefully weighted to balance the contributions from the signal, code, and feature levels, thus optimizing the performance as a whole:
\begin{equation}
\label{eq:l_overall}
\begin{split}
  \mathcal{L}_{overall} = \alpha\mathcal{L}_{phy} +\beta (\mathcal{L}_{code'}+\mathcal{L}_{code'}^{ppg}) +  \mathcal{L}_{distill},
\end{split}
\end{equation}
where we empirically set the hyperparameters $\alpha=2.0$, and $\beta=0.1$ respectively.
\section{Experimental Results}
\label{sec:experimets}

We comprehensively assess the performance of \textit{CodePhys} on four widely used benchmark datasets (VIPL-HR \cite{NiuVIPL2018}, UBFC-rPPG \cite{BobbiaUBFC2019}, PURE \cite{StrickerPURE2014}, and COHFACE \cite{HeuschCOHFACE2017}), in comparison with state-of-the-art methods. Experiments of rPPG-based heart rate (HR) measurement are conducted under both intra-dataset and cross-dataset settings. We further perform ablation studies to validate the effectiveness of our main technical designs. To demonstrate the effect of our key insight (i.e., code query process can uniformly mitigate visual interference), we evaluate the training effects of Stage I and attach the pre-trained codebook and decoder with different existing methods. Additionally, we assess the model's parameter size and inference speed. Finally, challenging cases of video quality degradation are tested to evaluate the robustness of our approach.

\subsection{Datasets and Performance Metrics}
\label{subsec:Datasets}

\textbf{VIPL-HR} serves as a large-scale dataset for remote physiological measurement under less-constrained scenarios, which contains 2,378 RGB videos of 107 subjects captured under diverse conditions, including various head movements, lighting conditions, and acquisition devices.  \textbf{UBFC-rPPG} includes 42 RGB videos recorded at 30 fps, captured under both sunlight and indoor illumination conditions. The ground-truth bio-signals are recorded by CMS50E with a 60 Hz sampling rate.  \textbf{PURE} is consisted of 60 RGB videos from 10 subjects, involving six different head motion tasks. These videos are recorded at 30 fps. Synchronized bio-signals are captured using CMS50E at a rate of 60 Hz. \textbf{COHFACE} contains 160 RGB videos from 40 subjects, with a frame rate of 20 fps. These videos are heavily compressed using MPEG-4 Visual, a factor noted in \cite{McDuffCOHFACEFactor2017} to potentially cause corruption of the rPPG signals.\\

\textbf{Performance metrics.} We perform HR estimation on all four datasets. Following \cite{NiuVIPL2018,LeeMeta2020}, we calculate mean absolute error (MAE), root mean square error (RMSE), standard deviation of the error (SD),  and Pearson's correlation coefficient ($r$) between the predicted HRs versus the ground-truth HRs.

\begin{table}[t]
\centering
\caption{Intra-dataset HR estimation testing results on VIPL-HR \cite{NiuVIPL2018}. The symbols $\triangle$, $\ddagger$, and $\star$ denote traditional, non-end-to-end learning based, and end-to-end learning based methods, respectively. The symbol $\downarrow$ indicates lower is better, and $\uparrow$ indicates higher is better. Best results are marked in \textbf{bold} and second best in \underline{underline}.}
\begin{tabular}{lcccc}
\toprule
Method & SD$\downarrow$ & MAE$\downarrow$ & RMSE$\downarrow$ & $r\uparrow$ \\
\midrule
SAMC\cite{TulyakovSAMC2016}$\triangle$ & 18.0 & 15.9 & 21.0 & 0.11 \\
CHROM\cite{DeCHROM2013}$\triangle$ & 15.1 & 11.4 & 16.9 & 0.28 \\
POS\cite{WangPOS2017}$\triangle$ & 15.3 & 11.5 & 17.2 & 0.30 \\
\midrule
RhythmNet\cite{NiuRhythm2020}$\ddagger$ & 8.11 & 5.30 & 8.14 & 0.76 \\
NAS-HR\cite{LuNASHR2021}$\ddagger$ & 8.10 & 5.12 & 8.01 & 0.79 \\
ST-Attention\cite{NiuSTAttn2019}$\ddagger$ & 7.99 & 5.40 & 7.99 & 0.66 \\
CVD\cite{NiuCVD2020}$\ddagger$ & 7.92 & 5.02 & 7.97 & 0.79 \\
Dual-GAN\cite{LuDualGAN2021}$\ddagger$ & 7.63 & 4.93 & 7.68 & 0.81 \\
ConDiff-rPPG\cite{WeiConDiff2024}$\ddagger$ & - & 4.81 & 7.76 & 0.81\\
NEST\cite{LuNEST2023}$\ddagger$ & 7.49 & 4.76 & 7.51 & \underline{0.84} \\
rPPG-HiBa\cite{HiBaLu2024}$\ddagger$ &\underline{7.26} & \underline{4.47} & \underline{7.28} & \textbf{0.85}\\

\midrule
DeepPhys\cite{ChenDeepPhys2018}$\star$ & 13.6 & 11.0 & 13.8 & 0.11 \\
PhysNet\cite{YuPhysNet2019}$\star$ & 14.9 & 10.8 & 14.8 & 0.20 \\
AutoHR\cite{YuAuto2020}$\star$ & 8.48 & 5.68 & 8.68 & 0.72 \\
PhysFormer\cite{YuPhysFormer2022}$\star$ & 7.74 & 4.97 & 7.79 & 0.78  \\
ContrastPhys+\cite{ContrastPhys+Sun2024}$\star$ & - & 7.49 & 14.4 & 0.49 \\
RS-rPPG\cite{RSrPPGZhao2024}$\star$ & - & 5.97 & 10.5 & 0.56 \\
DOHA\cite{DOHASun2023}$\star$ & 7.69 & 4.95 & 7.73 & 0.80 \\
\textbf{CodePhys(ours)}$\star$ & \textbf{7.07} & \textbf{4.27} & \textbf{7.11} & 0.81 \\
\bottomrule
\end{tabular}
\label{table:intra-VIPL}
\end{table}

\begin{table*}[t]
\centering
\caption{Intra-dataset HR estimation testing results on UBFC-rPPG \cite{BobbiaUBFC2019}, PURE \cite{StrickerPURE2014}, and COHFACE \cite{HeuschCOHFACE2017} datasets.}
\begin{tabular}{lccccccccc}
\toprule
\multirow{2}{*}{Method} & \multicolumn{3}{c}{UBFC-rPPG} & \multicolumn{3}{c}{PURE} & \multicolumn{3}{c}{COHFACE} \\
\cmidrule(lr){2-4} \cmidrule(lr){5-7} \cmidrule(lr){8-10}
& MAE$\downarrow$ & RMSE$\downarrow$ & $r\uparrow$& MAE$\downarrow$ & RMSE$\downarrow$& $r\uparrow$ & MAE$\downarrow$ & RMSE$\downarrow$ & $r\uparrow$\\
\midrule
Green\cite{VerkruysseGreen2008}$\triangle$ &7.50 &14.4 &0.62 &- &- &- &- &- &-\\
ICA\cite{PohICA2010}$\triangle$ &4.39 &11.6 &0.82 &5.70 &18.1 &0.70 &8.16 &13.9 &0.36\\
CHROM\cite{DeCHROM2013}$\triangle$ &3.10 &6.84 &0.93 &6.23 &17.1 &0.71 &8.44 &13.74 &0.34\\
POS\cite{WangPOS2017}$\triangle$ &3.52 &8.38 &0.90 &9.82 &13.4 &0.74 &6.58 &11.9 &0.49\\
\midrule
RhythmNet\cite{NiuRhythm2020}$\ddagger$ &5.59 &6.82 &0.72 &2.71 &4.86 &0.98 &- &- &-\\
Dual-GAN\cite{LuDualGAN2021}$\ddagger$ &0.44 &\underline{0.67} &\underline{0.99} &{0.82} &\underline{1.31}&\underline{0.99} &- &- &-\\
\midrule
HR-CNN\cite{SpetlikHRCNN2018}$\star$ & - & - & - & 1.80 & 2.40 & 0.98 & 8.10 & 10.8 & 0.29 \\
ETA-rPPGNet\cite{hu2021eta}$\star$ &- &- &- &- &- &- &4.67 &6.65 &0.77\\
AND-rPPG\cite{BirlaAnd2022}$\star$ & 2.67 & 4.07 & 0.96 & - & - & - & 3.82 & 5.10 & 0.79 \\
EfficientPhys\cite{LiuEfficientPhys2021}$\star$ & 1.14 &1.81 &0.99 &1.33& 5.99& 0.97& - &-& -\\
TranPhys\cite{shao2023tranphys}$\star$ &4.66 &7.24 &0.85 &- &- &- &5.04 &7.36 &0.83\\
PhysNet\cite{YuPhysNet2019}$\star$ & 2.38 & 3.19 & - & 2.16 & 2.70 & - & 5.38 & 10.8 & - \\
Gideon et al.\cite{GideonHeart2021}$\star$ & 1.85 & 4.28 & 0.93 & 2.30 & 2.90 & 0.99 & 1.50 & 4.60 & \underline{0.90} \\
PulseGAN\cite{SongPulseGAN2021}$\star$ & 1.19 & 2.10 & 0.98 & 2.28 & 4.29 & 0.99 & - & - & -\\
Contrast-Phys\cite{SunContrastPhys2022}$\star$ &0.64 &1.00 &0.99 &1.00 &1.40 &0.99 &- &- &-\\
Face2PPG\cite{AlvarezFace2PPG2023}$\star$&0.80 & 1.10 & - & 1.20 & 1.80 & - & 7.50 & 9.80\\
ContrastPhys+\cite{ContrastPhys+Sun2024}$\star$ & 0.64 & {1.00} & 0.99 & 1.00 & 1.40 & 0.99 &- &- &-\\
SiNC\cite{SiNCSpeth2023}$\star$ & {0.59} & 1.83 & 0.99 & \underline{0.61} & 1.84 & 0.99&- &- &-\\
DeeprPPG\cite{liu2020Deeprppg}$\star$ &- &- &- &- &- &- &3.07 &7.86 &0.86\\
Li et al.\cite{LiMotionRobust2023}$\star$ &0.76 &1.62 &- &1.44 &2.50 &- &\underline{1.31} &\underline{3.92} &-\\
DOHA\cite{DOHASun2023}$\star$ & 1.41 & 1.56 & 0.98 & 0.95 & 1.58 & 0.99&- &- &-\\
PhysFormer\cite{YuPhysFormer2022}$\star$ & \underline{0.40} & 0.71 & {0.99} & 1.10 & 1.75 & {0.99} & - & - & - \\
Yue et al.\cite{YueVideo2023}$\star$ &0.58 &0.94 &0.99 &1.23 &2.01 &0.99 &- &- &-\\

\textbf{CodePhys(ours)}$\star$ & \textbf{0.21} & \textbf{0.26} & \textbf{0.99}& \textbf{0.39} & \textbf{0.83} & \textbf{0.99}& \textbf{1.19} & \textbf{2.75} & \textbf{0.97} \\
\bottomrule
\end{tabular}
\label{table:intra-UCP}
\end{table*}

\begin{figure}[t]
\centering
\includegraphics[width=1\columnwidth]{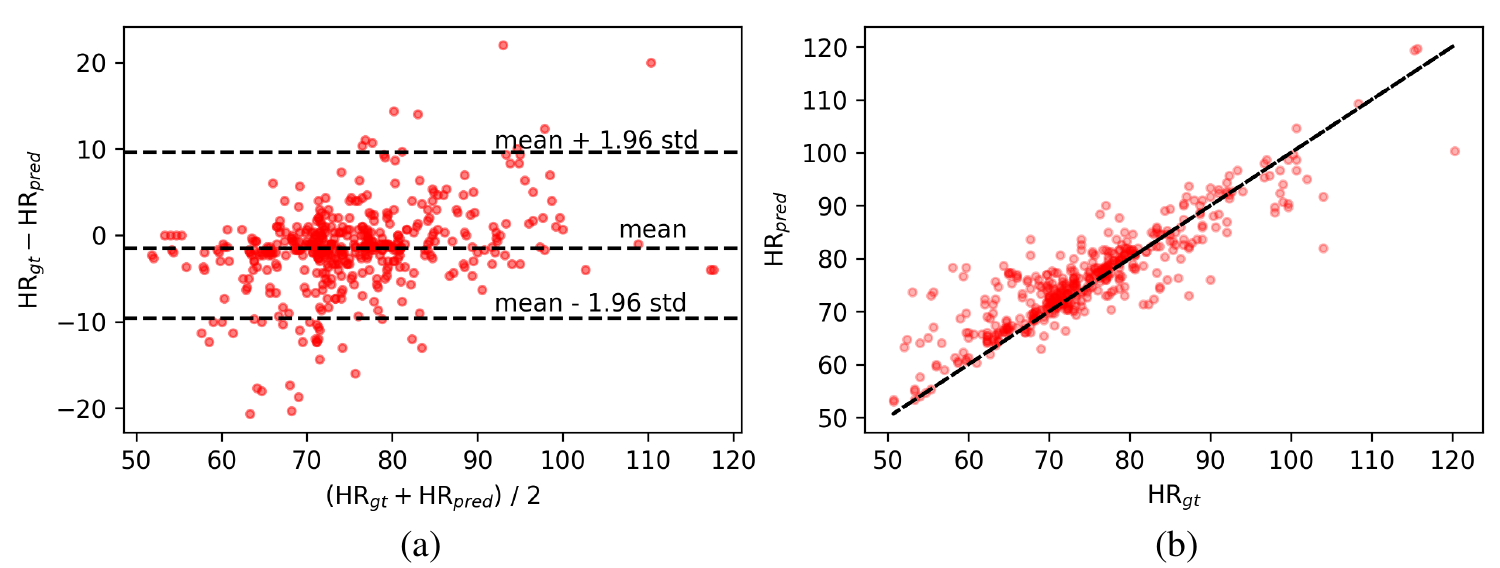}
\caption{The Bland-Altman plot (a) and scatter plot (b) show the difference between  ground-truth HRs and predicted HRs by CodePhys on the Fold-1 of VIPL-HR dataset.}
\label{fig:bland_altman}
\end{figure}

\begin{figure}[t]
\centering
\includegraphics[width=1\columnwidth]{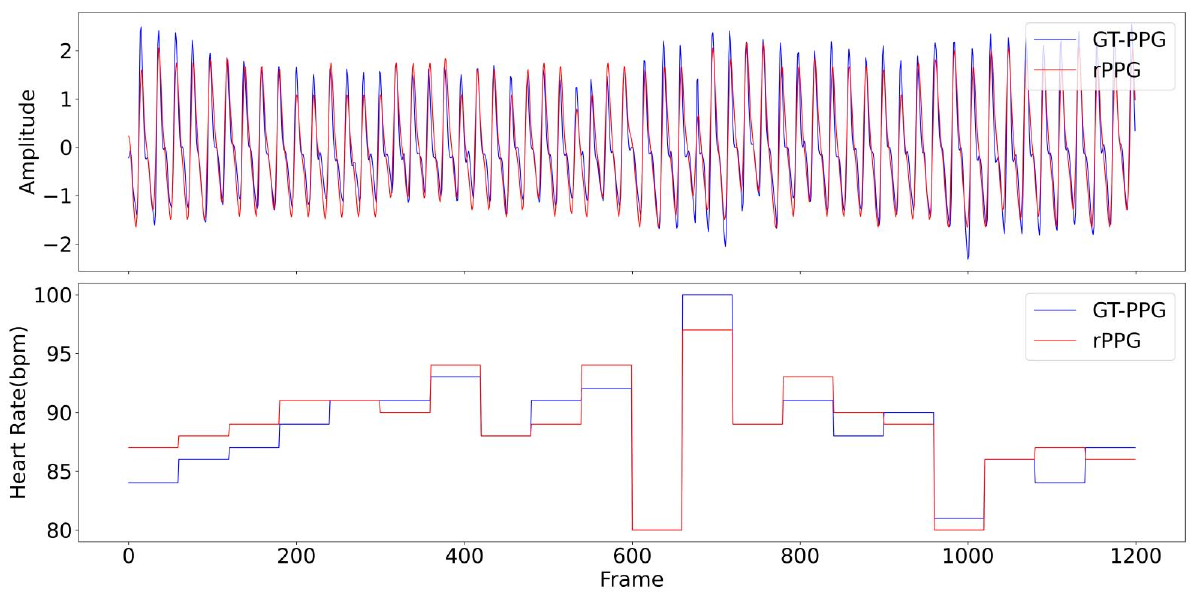}
\caption{Visual comparison of the rPPG signals (top) and HRs (bottom) predicted by CodePhys on UBFC-rPPG dataset, alongside the corresponding ground-truth.}
\label{fig:rppg_hr}
\end{figure}

\begin{figure}[t]
\centering
\includegraphics[width=1\columnwidth]{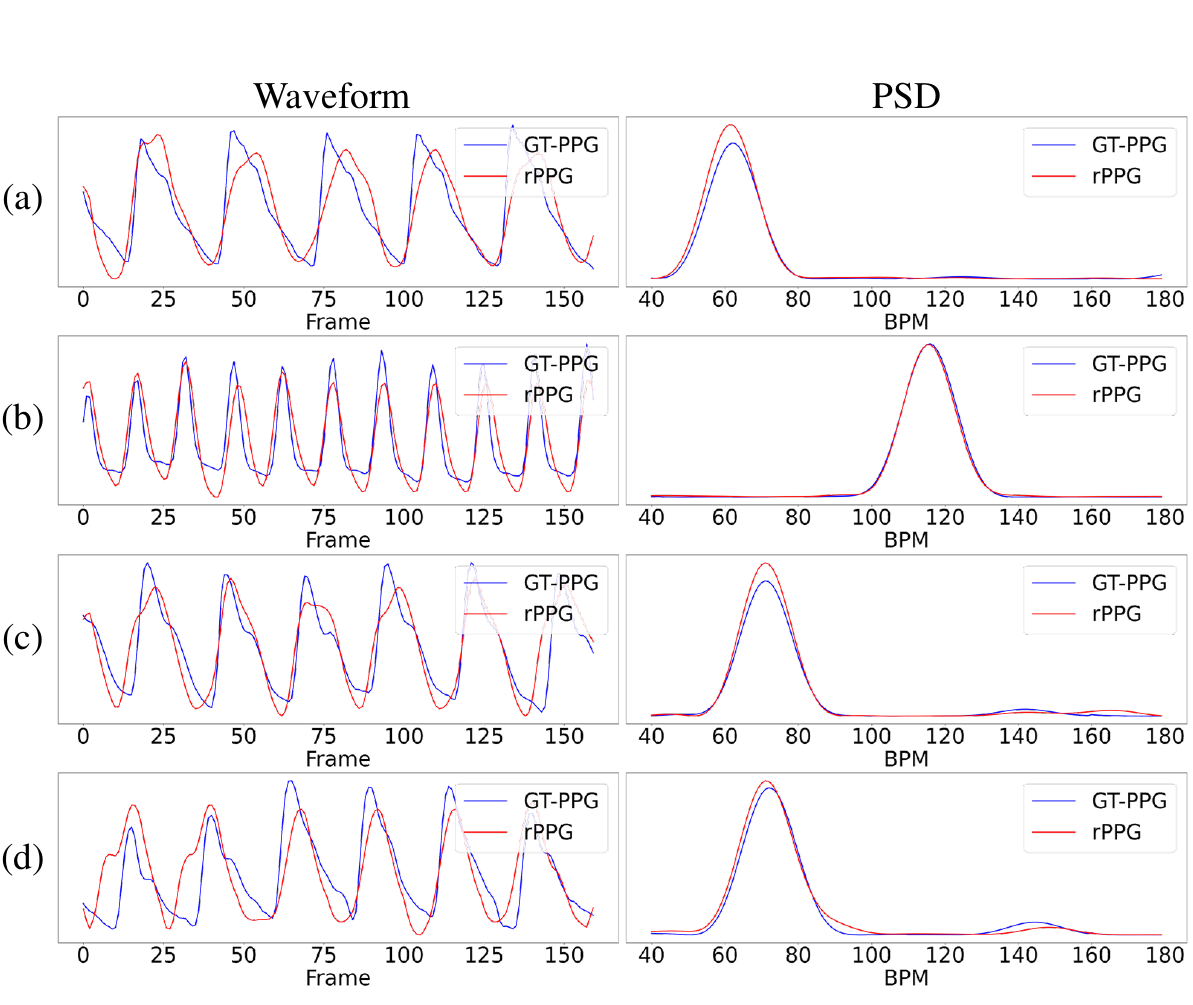}
\caption{Visual comparison of the rPPG signals (left) predicted by CodePhys and their corresponding PSDs (right), alongside the respective ground-truth. (a) VIPL-HR, (b) UBFC-rPPG, (c) PURE, and (d) COHFACE.}
\label{fig:rppg_psd}
\end{figure}
\subsection{Implementation Details}
\label{subsec:implementation}

\textbf{Pre-process settings.} Our proposed method is implemented with PyTorch. Following \cite{LuNEST2023}, we use the FAN \cite{BulatFAN2017} face detector to find the landmarks for each frame, then we crop and align the video frames to 128$\times$128 pixels according to the obtained landmarks. We uniformly adjust the video frame rate to 30 fps for efficiency.

\textbf{Train settings.} In Stage I, we train signal autoencoder network for 15 epochs using the Adam \cite{KingmaADAM2015} optimizer, where the learning rate is initialized to 5e-3, and the batch size is set to 8. In Stage II, we use the Adam optimizer with a batch size of 4 for 20 epochs. The base learning rate and weight decay are set to 1e-4 and 5e-5, respectively. Both training stages are conducted on one NVIDIA GeForce RTX3090 GPU.

\textbf{Inference settings.} Following \cite{NiuRhythm2020, YuPhysFormer2022}, we split the 30-second video into three 10-second clips in the testing stage. We calculate HRs for these clips and then average them to obtain the video-level HR. Note that the datasets are uniformly divided throughout both stages.

\subsection{Intra-dataset Testing}
\label{subsec:intra-data_eval}

We first evaluate the HR estimation on all datasets under intra-dataset setting. We compare our method with 32 methods, including traditional methods \cite{PohICA2010,DeCHROM2013,WangPOS2017,TulyakovSAMC2016,VerkruysseGreen2008}, non-end-to-end methods \cite{LuDualGAN2021,NiuRhythm2020,LuNASHR2021,NiuCVD2020,LuNEST2023,NiuSTAttn2019,WeiConDiff2024, HiBaLu2024}, and end-to-end methods \cite{YuAuto2020,SpetlikHRCNN2018,BirlaAnd2022,ChenDeepPhys2018,YuPhysNet2019,GideonHeart2021,SongPulseGAN2021,YuPhysFormer2022,shao2023tranphys,YueVideo2023,SunContrastPhys2022,LiMotionRobust2023,liu2020Deeprppg,hu2021eta,AlvarezFace2PPG2023,ContrastPhys+Sun2024,SiNCSpeth2023,DOHASun2023,RSrPPGZhao2024}. TABLE \ref{table:intra-VIPL} displays intra-dataset testing results for VIPL-HR, while TABLE \ref{table:intra-UCP} shows the results for UBFC-rPPG, PURE, and COHFACE.

Note that our CodePhys adopts a hybrid approach that incorporates extensive pre-training in Stage I, followed by an end-to-end process in Stage II. Unlike typical non-end-to-end STMap-based approaches that rely heavily on the hand-crafted feature pre-processing, CodePhys performs all processing steps without external hand-crafted feature extraction. Besides, if we have obtained the pre-trained noise-free PPG representations in the codebook, we only need to conduct the end-to-end process (Stage II) in the training phase. Therefore, we categorize the CodePhys as an end-to-end neural network process.


\subsubsection{HR estimation on VIPL-HR}  Following the protocol in \cite{NiuRhythm2020}, we employ a subject-exclusive 5-fold cross-validation protocol on VIPL-HR. We compare CodePhys with three traditional methods, eight non-end-to-end methods, and seven end-to-end methods on VIPL-HR. As shown in TABLE \ref{table:intra-VIPL}, the proposed CodePhys achieves the lowest SD (7.07 bpm), MAE (4.27  bpm), and RMSE (7.11 bpm) among all the methods, outperforming previous methods by a large margin. 
We visualize the Bland-Altman plot and scatter plot on the Fold-1 of VIPL-HR dataset in Fig. \ref{fig:bland_altman}. HR$_{pred}$ and HR$_{gt}$ denote the HRs computed from the predicted rPPG signals and corresponding GT-PPG signals, respectively. Each point represents the estimation results of a test sample. The x-axis in the Bland-Altman plot indicates the average of HR$_{gt}$ and HR$_{pred}$, while the y-axis represents their differences. The top and bottom dashed lines indicate the confidence intervals for the 95\% limits of agreement. It can be observed that HR$_{gt}$ and HR$_{pred}$ have a good correlation over a wide range from 50 bpm to 120 bpm. CodePhys achieves a better MAE than the second best method (4.27 bpm for CodePhys and 4.47 bpm for rPPG-HiBa \cite{HiBaLu2024}) on VIPL-HR. Considering that VIPL-HR is a large dataset collected in less-constrained scenarios, the results demonstrate the robustness of CodePhys against visual disturbances such as head movements and occlusions.

\subsubsection{HR estimation on UBFC-rPPG} On the UBFC-rPPG dataset, we use the videos of the first 30 subjects for training and the videos of the remaining 12 subjects for testing according to \cite{SongPulseGAN2021}. The HR estimation results are shown in TABLE \ref{table:intra-UCP}. The proposed CodePhys outperforms the existing state-of-the-art methods on MAE (0.21 bpm), RMSE (0.26 bpm), and $r$ (0.99) metrics for HR prediction. We randomly select a video sample and plot the estimated rPPG signal and short-term HR changes in Fig. \ref{fig:rppg_hr}. The results show that CodePhys not only fits the rPPG signal consistently but also tracks HR changes effectively over a long period.

\subsubsection{HR estimation on PURE} Following the protocol in \cite{LuDualGAN2021}, we compare CodePhys with 23 methods. It can be seen from TABLE \ref{table:intra-UCP} that the proposed CodePhys achieves the best performance on all evaluation metrics for HR, especially compared with the second best method SiNC \cite{SiNCSpeth2023}. CodePhys achieves an improvement of 0.22 bpm in MAE (compared to 0.61 bpm) and an improvement of 1.01 bpm in RMSE (compared to 1.84 bpm), which is a significant improvement. These results demonstrate the robustness of CodePhys against the interference induced by various head movements.

\subsubsection{HR estimation on COHFACE} The facial videos in the COHFACE dataset are highly compressed, which degrades the video quality. As illustrated in TABLE \ref{table:intra-UCP}, CodePhys still achieves state-of-the-art performance across all evaluation metrics. These results show that CodePhys maintains strong performance even under highly compressed conditions.

We randomly select clip samples from each of the four datasets and plot the predicted rPPG and the corresponding PSD signals in Fig. \ref{fig:rppg_psd}. The results clearly demonstrate that CodePhys effectively predicts the rPPG signals across different datasets.
\begin{table}[t]
\centering
\caption{Cross-dataset HR estimation testing results on UBFC-rPPG, PURE, and COHFACE datasets.}
\setlength{\tabcolsep}{0.4mm}{
\begin{tabular}{lcccccc}
\toprule
\multirow{2}{*}{Method} & \multicolumn{2}{c}{P+C$\rightarrow$U}& \multicolumn{2}{c}{U+C$\rightarrow$P} & \multicolumn{2}{c}{U+P$\rightarrow$C}\\
\cmidrule(lr){2-3} \cmidrule(lr){4-5} \cmidrule(lr){6-7}
& MAE$\downarrow$ & RMSE$\downarrow$ & MAE$\downarrow$ & RMSE$\downarrow$ & MAE$\downarrow$ & RMSE$\downarrow$\\
\midrule
GREEN\cite{VerkruysseGreen2008}$\triangle$ & 8.29 & 15.8 & 9.03 & 13.9 & 10.9 & 16.7 \\
ICA\cite{PohICA2010}$\triangle$ & 4.39 & 11.6 & 15.2 & 21.3 & 14.3 & 19.3 \\
POS\cite{WangPOS2017}$\triangle$ & 3.52 & 8.38 & 22.3 & 30.2 & 19.9 & 24.6 \\
CHROM\cite{DeCHROM2013}$\triangle$ & 3.10 & 6.84 & 3.82 & 6.80 & 7.80 & 12.5 \\
\midrule
Dual-GAN\cite{LuDualGAN2021}$\ddagger$ & 0.74 & \underline{1.02} & - & - & - & -\\
\midrule
Multi-task\cite{TsouMulti2020}$\star$ & 1.06 & 2.70 & 4.24 & 6.44 & - & -\\
DG-rPPGNet\cite{ChungDomain2022}$\star$ & \underline{0.63} & 1.35 & \underline{3.02} & \underline{4.69} & \underline{7.19} & \underline{8.99}\\
\textbf{CodePhys(ours)}$\star$ & \textbf{0.58} & \textbf{0.96} & \textbf{0.70} & \textbf{1.39} & \textbf{6.04} & \textbf{8.71} \\
\bottomrule
\end{tabular}}
\label{table:cross-UPC}
\end{table}

\begin{table}[t]
\centering
\caption{Cross-dataset HR estimation testing results on PURE dataset.}
\begin{tabular}{lcccc}
\toprule
Method & SD$\downarrow$ & MAE$\downarrow$ & RMSE$\downarrow$ & $r\uparrow$ \\
\midrule
ICA\cite{PohICA2010}$\triangle$ &19.4 &16.1 &20.9 &0.41 \\
CHROM\cite{DeCHROM2013}$\triangle$ &14.8 &11.1 &15.9 &0.52\\
POS\cite{WangPOS2017}$\triangle$ & 11.2& 10.9& 13.4 &0.63 \\
\midrule
RhythmNet\cite{NiuRhythm2020}$\ddagger$ & 11.7 &9.09 &11.9 &- \\
\midrule
PhysNet\cite{YuPhysNet2019}$\star$ & 11.5& 8.72 &10.4 &0.69 \\
Meta-rPPG\cite{LeeMeta2020}$\star$ & 9.31& 6.94& 10.1 &0.71\\
PhysFormer\cite{YuPhysFormer2022}$\star$ & 9.05 &6.13& 9.18& \underline{0.82} \\
TranPhys\cite{shao2023tranphys}$\star$&\underline{8.03} &\underline{5.31} &\underline{8.41}& 0.76 \\
\textbf{CodePhys(ours)}$\star$ & \textbf{6.54} & \textbf{4.03} & \textbf{7.02} & \textbf{0.83} \\
\bottomrule
\end{tabular}
\label{table:cross-VP}
\end{table}

\subsection{Cross-dataset Testing}
\label{subsec:cross-dataset_eval}

We perform cross-dataset testing across UBFC-rPPG, PURE, and COHFACE datasets to assess the generalization capability of CodePhys. Following the testing protocol of \cite{ChungDomain2022}, we train CodePhys on two datasets and test on the remaining one. For example, P+C$\rightarrow$U represents training on PURE and COHFACE datasets while testing on UBFC-rPPG dataset. The cross-dataset HR estimation results of our method and baseline methods are presented in TABLE \ref{table:cross-UPC}. Following the cross-dataset testing protocol of TranPhys \cite{shao2023tranphys}, we further evaluate the generalization of CodePhys trained on VIPL-HR dataset. TABLE \ref{table:cross-VP} shows the corresponding results obtained by training on the VIPL-HR dataset and testing on PURE dataset. CodePhys outperforms state-of-the-art methods under all evaluation metrics. These results suggest that incorporating a noise-free codebook with prior knowledge significantly improves the generalization ability.

\subsection{Ablation Study}
\label{subsec:ablation}

\begin{table}[t]
\caption{Ablation study of the main components of CodePhys on VIPL-HR dataset.}
\centering
\begin{tabular}{ccccccccc}
\toprule
CQP & APB & SFD & SAM & Stage I & MAE$\downarrow$ & RMSE$\downarrow$ & $r\uparrow$ \\
\midrule
  & $\checkmark$& $\checkmark$&$\checkmark$  & $\checkmark$ & 5.06 & 8.83 & 0.67\\
$\checkmark$ & & $\checkmark$&$\checkmark$ & $\checkmark$ & 4.89 & 8.82 & 0.67\\
$\checkmark$ & $\checkmark$&  &$\checkmark$  & $\checkmark$ & 4.59 & 8.06 & 0.73\\
$\checkmark$ & $\checkmark$& $\checkmark$&   & $\checkmark$ & 4.50 & 7.94 & 0.75\\
$\checkmark$ & $\checkmark$& $\checkmark$& $\checkmark$ & &5.27&9.15&0.65\\
$\checkmark$ & $\checkmark$& $\checkmark$&$\checkmark$ & $\checkmark$ & \textbf{4.30} & \textbf{7.29} & \textbf{0.80}\\
\bottomrule
\end{tabular}
\label{table:ablation_all}
\end{table}

We perform ablation studies on the key components of CodePhys, including the code query process (CQP), auxiliary prior branch (APB), soft feature distillation (SFD), spatial attention mechanism (SAM), and the pre-trained signal autoencoder (Stage I) on the Fold-1 of VIPL-HR dataset. Additionally, we conduct sensitivity experiments on the hyperparameters within CodePhys.

\begin{table}[t]
\centering
\caption{Ablation study of the code query process on PURE and COHFACE datasets.}
\begin{tabular}{lccccc}
\toprule
Dataset& CQP& SD$\downarrow$ & MAE$\downarrow$ & RMSE$\downarrow$ & $r\uparrow$ \\
\midrule
\multirow{2}{*}{PURE\cite{StrickerPURE2014}} && 3.31& 1.22 & 3.44 & 0.96\\
& $\checkmark$&\textbf{0.82}& \textbf{0.39} &\textbf{0.83}& \textbf{0.99}\\
\midrule
\multirow{2}{*}{COHFACE\cite{HeuschCOHFACE2017}}& & 3.81& 1.42& 3.93 &0.94\\
& $\checkmark$ &\textbf{2.71}& \textbf{1.19} &\textbf{2.75} &\textbf{0.97}\\
\bottomrule
\end{tabular}
\label{table:ablation_CQP_PC}
\end{table}

\begin{table}[t]
\caption{PPG reconstruction results on benchmark datasets.}
\centering
\begin{tabular}{lcccc}
\toprule
Dataset & SD$\downarrow$ & MAE$\downarrow$ & RMSE$\downarrow$ & $r\uparrow$ \\

\midrule
VIPL-HR\cite{NiuVIPL2018} & 0.19 & 0.05 & 0.19 & 0.99 \\
UBFC-rPPG\cite{BobbiaUBFC2019} & 0.20 & 0.13 & 0.20 & 0.99 \\
PURE\cite{StrickerPURE2014} & 0.21 & 0.13 & 0.21 & 0.99 \\
COHFACE\cite{HeuschCOHFACE2017} & 0.09 & 0.02 & 0.09 & 0.99 \\
\bottomrule
\end{tabular}\vspace{-0.5em}
\label{table:ablation_CQP_S1}
\end{table}

\begin{figure}[t]
\centering
\includegraphics[width=1\columnwidth]{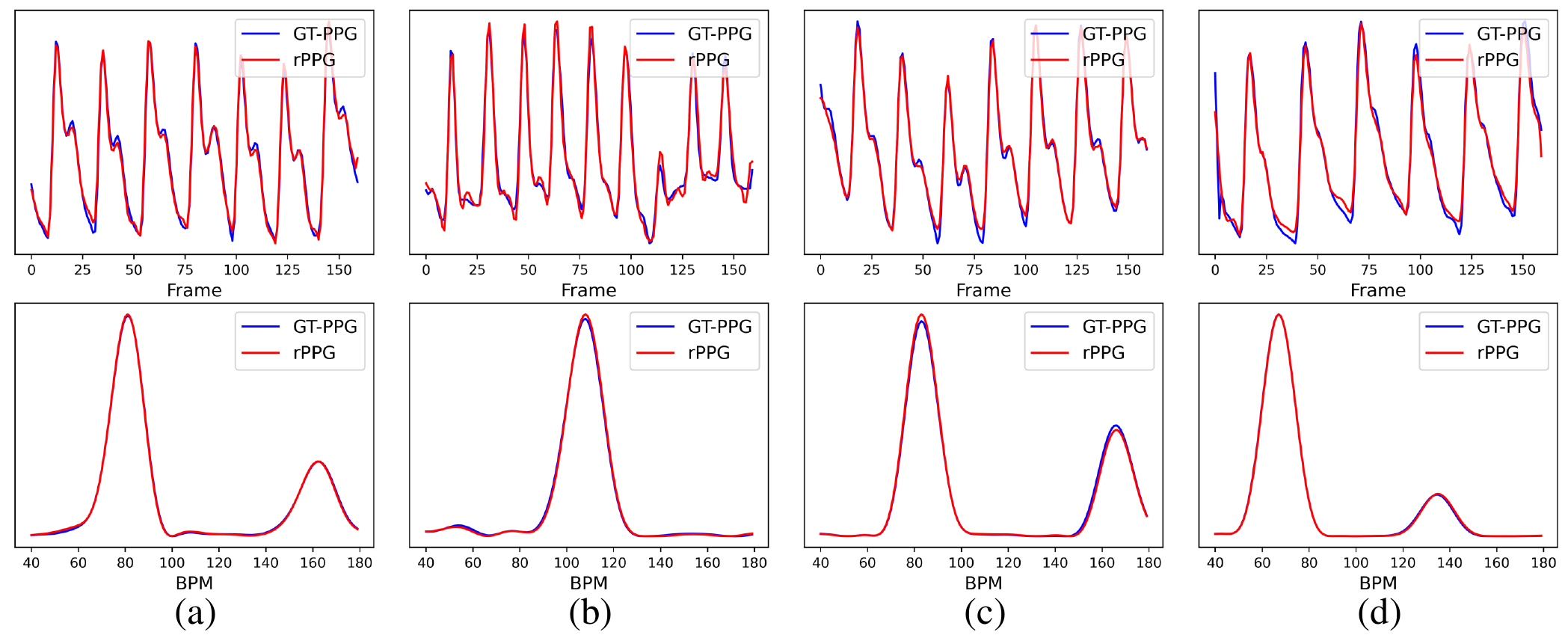}
\caption{Visual comparison of rPPG signals (top) reconstructed by the signal autoencoder network and their corresponding PSDs (bottom), alongside the corresponding ground-truth. (a) VIPL-HR, (b) UBFC-rPPG, (c) PURE, and (d) COHFACE datasets.}
\label{fig:ablation_CQP_S1}
\end{figure}

\subsubsection{Code query process (CQP)} CQP is the process of selecting noise-free PPG features from the codebook $\mathbf{C}$ to replace noisy rPPG features $\mathbf{Z}_{rppg}$. The process aims to achieve noise-free rPPG features $\hat{\mathbf{Z}}_{rppg}$, which is the key process to mitigate visual interference. To validate the effectiveness of the code query process, we construct a comparative experiment designed to remove CQP. Specifically, we directly extract the rPPG signal from $\mathbf{Z}_{rppg}$ instead of the noise-free version $\hat{\mathbf{Z}}_{rppg}$. The rPPG signal could be obtained by $\boldsymbol{s}_{pred}=D_s(\mathbf{Z}_{rppg})$. The results of comparative experiment are shown in the first row of TABLE \ref{table:ablation_all}. CQP reduces the MAE by 0.76 bpm, and the RMSE by 1.54 bpm while increasing the correlation coefficient r by 0.13. We additionally conduct ablation experiments on CQP in PURE and COHFACE datasets, with results shown in TABLE \ref{table:ablation_CQP_PC}. These results clearly indicate that CQP contributes to the improvement of the rPPG estimation in datasets with substantial head motion, occlusions, and video compression.

Furthermore, the ability of CQP to eliminate the visual interference depends on the noise-free PPG features learned in Stage I. To evaluate the effectiveness of learned noise-free codebook, we present the reconstruction results on four datasets in TABLE \ref{table:ablation_CQP_S1} and illustrate the reconstruction of several signal samples from each dataset in Fig. \ref{fig:ablation_CQP_S1}. The results clearly demonstrate that CodePhys achieves nearly error-free reconstruction in Stage I, indicating the signal autoencoder network can sample from the noise-free proxy space and recover corresponding PPG signals. This capability is crucial for the effectiveness of CQP in reducing visual interference during Stage II. 

 


\subsubsection{Auxiliary prior branch (APB)} 
In this ablation study, we simply remove the auxiliary information from APB to evaluate its impact on the model. The comparison between the second and sixth rows in TABLE \ref{table:ablation_all} clearly demonstrates that the distribution knowledge embedded in the pre-trained signal encoder is effective.

\subsubsection{Soft feature distillation (SFD)} To evaluate the effectiveness of soft feature distillation, we conduct an ablation study by removing the soft feature distillation loss $\mathcal{L}_{distill}$ during  Stage II. The results are shown in the third row of TABLE \ref{table:ablation_all}. SFD brings a significant increase in MAE (0.29 bpm) and RMSE (0.77 bpm) and a decrease in $r$ (0.07). These results underscore the importance of SFD in facilitating the extraction of quasi-periodic video features.


\begin{figure}[t]
\centering
\includegraphics[width=1\columnwidth]{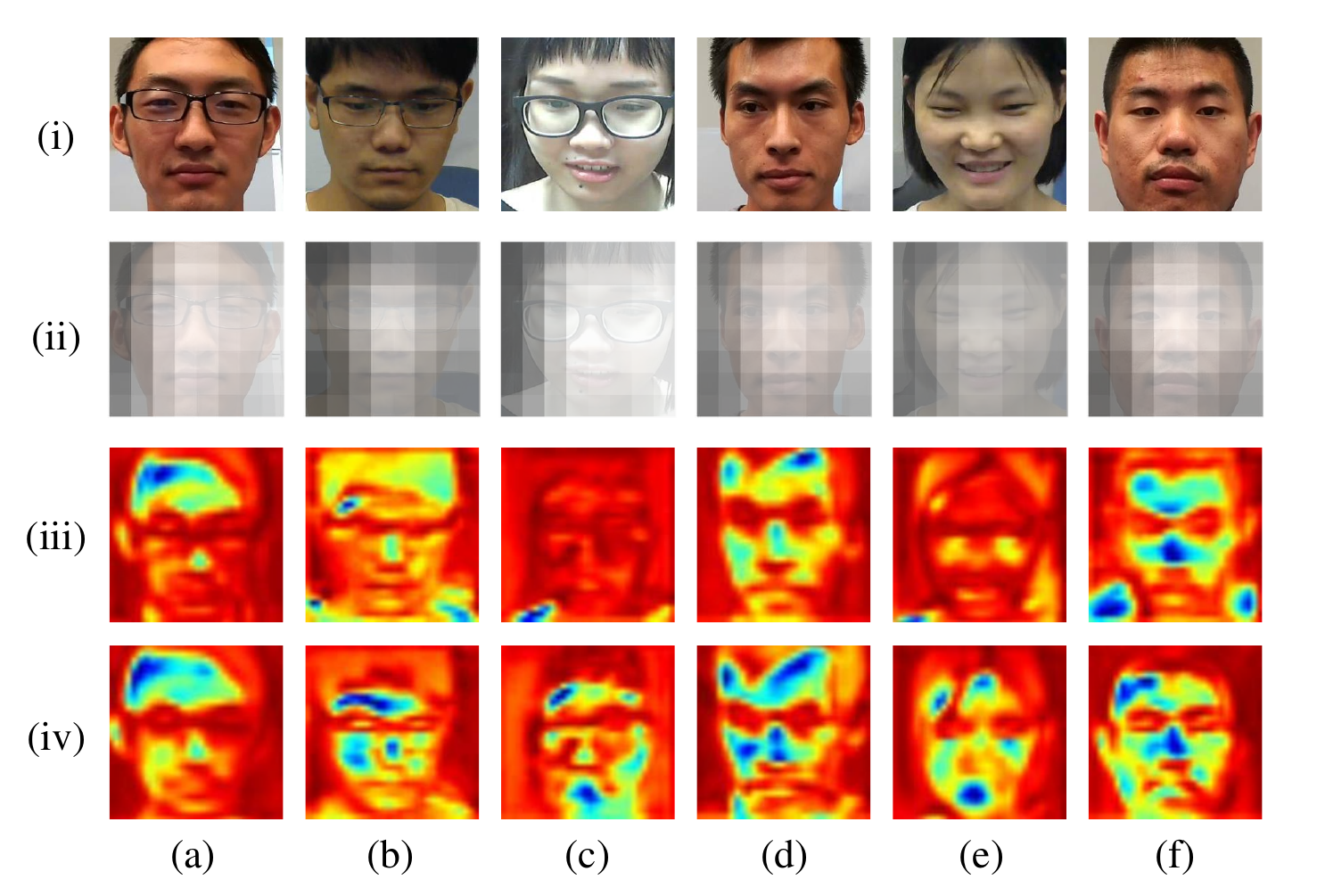}
\caption{Visualization of spatial attention maps and the corresponding video features. The rows (i), (ii), (iii), and (iv) denote the original facial images, the attention maps $\mathbf{W}_v$ predicted by CodePhys with SAM (the brighter, the more attentive), the feature maps $\mathbf{F}_v$ predicted by CodePhys before SAM, and the feature maps $\mathbf{F}_v^{sa}$ predicted by CodePhys after SAM (blue color indicates more attentive), respectively. The results (a)-(f) are randomly selected from the VIPL-HR dataset.}
\label{fig:albation_SAM}
\end{figure}

\subsubsection{Spatial attention mechanism (SAM)} To investigate the effectiveness of SAM, we conduct an ablation study by removing SAM from the video feature extractor. Specifically, the output of video feature extractor $E_{v}$ is $\mathbf{F}_v$, rather than the spatial-aware video feature $\mathbf{F}_{v}^{sa}$. The results of this ablation study are shown in the fourth row of TABLE \ref{table:ablation_all}, which show that the absence of SAM leads to an increase of 0.20 bpm in MAE, and 0.65 bpm in RMSE. The results suggest that SAM can adaptively identify informative regions of physiological reflection, thus enhancing overall performance. Meanwhile, to verify whether SAM correctly captures facial information, we randomly select and visualize six samples in VIPL-HR and present the results in Fig. \ref{fig:albation_SAM}. It can be seen that SAM eliminates some incorrect attention predictions (e.g., samples
b and f incorrectly focus on the hair, collar, and background areas), which could help to achieve more reasonable facial ROIs on physiological feature extractions. These results clearly show that SAM effectively captures facial physiological features, which significantly aids in boosting the model's performance.

\subsubsection{Pre-trained signal autoencoder network (Stage I)} In Stage I, CodePhys trains the signal autoencoder network and codebook $\mathbf{C}$ for noise-free PPG representation. To verify the necessity of Stage I, we train Stage II directly without the pre-trained weights. The results of the experiment are shown in the fifth row of TABLE \ref{table:ablation_all}. The experiment without pre-trained weights brings a significant increase in MAE (0.97 bpm) and RMSE (1.86 bpm), and a substantial decrease in $r$ (0.15), indicating that the pre-training in Stage I holds important position. 

\begin{figure*}[t]
\centering
\includegraphics[width=1.0\linewidth]{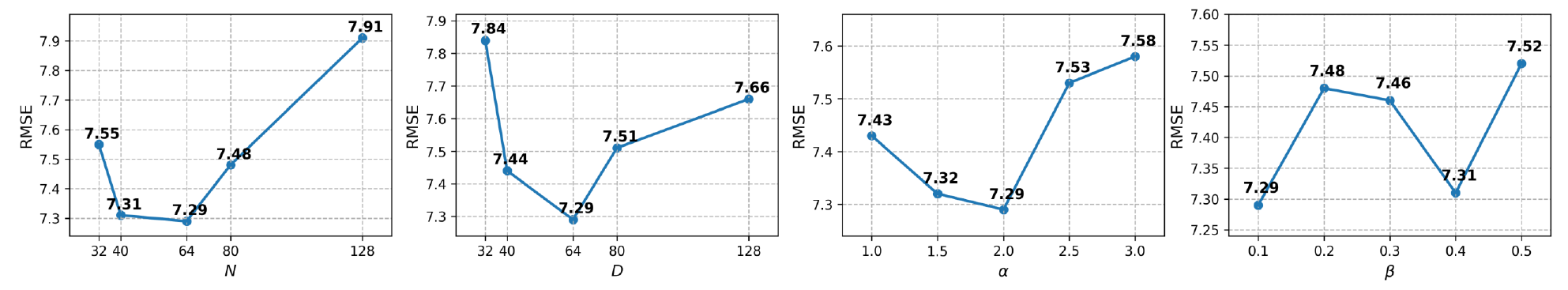 } 
\caption{Ablation study of the hyperparameters  of $N$, $D$ in the codebook $\mathbf{C}$, and $\alpha$, $\beta$ in the loss function $\mathcal{L}_{overall}$ (from left to right) . }
\label{fig:hpyerparams}
\end{figure*}

\subsubsection{Hyperparameters in CodePhys} We have described all the configurations in section \ref{sec:method}, including the codebook settings and the weights trade-off in the loss function. These hyperparameters are selected based on experience and the scale of the loss terms, but the performance of CodePhys is not sensitive to them. To verify that our chosen configuration for CodePhys is suitable, we conduct ablation studies on the number of PPG features stored in the codebook ($N$), the dimension of each PPG feature ($D$), and the parameters $\alpha$, $\beta$ in the loss function $\mathcal{L}_{overall}$. Note that only one parameter is changed at one time, while the others remain unchanged.

Fig. \ref{fig:hpyerparams} presents the results of the ablation study on hyperparameters in CodePhys. As we discussed earlier, it is clear that selecting an appropriate number $N$ of PPG features and the dimension $D$ of each PPG feature could benefit the experimental results, while both abundant and insufficient capacity can lead to slight performance degradation. Meanwhile, it can be observed that variations in $\alpha$ and $\beta$ within a certain range do not significantly affect the performance of CodePhys.

\subsection{Efficiency Analysis}
\label{subsec:effiency_analysis}
\begin{table}[t]
\caption{Comparison of parameters and computational cost.}
\centering
\begin{tabular}{lcccc}
\toprule
 \multirow{2}{*}{Method} & \#Param. & MACs & TIME $\downarrow$& RMSE $\downarrow$ \\
         & (M) & (G)& (ms) & (bpm)\\        
 
\midrule
PhysNet\cite{YuPhysNet2019} & 0.73 & 70.12 & 0.14 & 14.8 \\
TS-CAN\cite{LiuTSCAN2020} & 5.26 & 162.2 & 0.77 & 14.2 \\
DeepPhys\cite{ChenDeepPhys2018} & 7.51 & 135.5 & 0.67 & 13.8 \\
AutoHR\cite{YuAuto2020} & 0.99 & 189.2 & - & 8.68 \\
PhysFormer\cite{YuPhysFormer2022} & 7.03 & 47.01 & 0.18 & 7.79 \\
PhysFormer++\cite{YuPhysformer++2023} & 9.79 & 51.76 & 0.25 & 7.62 \\
\textbf{CodePhys(ours)} & 5.73 & 75.79 & \textbf{0.12} & \textbf{7.07} \\
\bottomrule
\end{tabular}
\label{table:flops}
\end{table}

The paradigm of CodePhys includes extra codebook and decoder modules compared to most existing methods. To verify whether the superior performance of CodePhys comes from a larger model capacity, we compare computational cost with some recent end-to-end frameworks on the VIPL-HR dataset. The number of parameters, multiply–accumulates (MACs), and the inference time per frame (TIME) are shown in TABLE \ref{table:flops}. The MACs and inference time per frame are calculated with the video input size 3$\times$160$\times$128$\times$128 ($C\times T\times H\times W$) on one RTX3090 GPU for all frameworks. CodePhys exhibits the best RMSE and the least inference time with reasonable training efficiency.

\subsection{Integrating existing methods into CodePhys framework}
\label{subsec:impl_other_methods}

\begin{table}[t]
\caption{Study of integrating existing methods into CodePhys framework.}
\centering
\begin{tabular}{lcccc}
\toprule
\multirow{2}{*}{Method} & CodePhys& \multirow{2}{*}{MAE$\downarrow$} & \multirow{2}{*}{RMSE$\downarrow$} & \multirow{2}{*}{$r\uparrow$}\\

     & Framework & & & \\
\midrule
\multirow{2}{*}{DeepPhys\cite{ChenDeepPhys2018}} & w/o & 11.0 & 13.8 & 0.11 \\ 
 & w. &6.71 & 11.1 & 0.47 \\ 
\midrule
\multirow{2}{*}{EfficientPhys\cite{LiuEfficientPhys2021}} & w/o &7.17 & 11.8 & 0.43 \\ 
 & w. &6.25 & 10.8 & 0.52\\ 
\midrule
\multirow{2}{*}{PhysFormer\cite{YuPhysFormer2022}}& w/o &4.97 & 7.79 & 0.78 \\ 
& w. &4.62 & 7.63 & 0.79\\ 
\bottomrule
\end{tabular}
\label{table:other_impl}
\end{table}

CodePhys can serve as a plug-and-play framework to improve the robustness of existing end-to-end methods. To validate the effectiveness of the paradigm, we choose some recent CNN-based and transformer-based end-to-end frameworks (i.e., DeepPhys \cite{ChenDeepPhys2018}, EfficientPhys \cite{LiuEfficientPhys2021}, and PhysFormer \cite{YuPhysFormer2022}) as backbone networks. As shown in Fig. \ref{fig:concept}(a), most end-to-end networks could be divided into two parts (i.e., the video feature extractor and the rPPG estimator). We adaptively replace the query feature extractor in Fig. \ref{fig:concept}(c) with the video feature extractor part of these backbone networks, thereby achieving the integration of CodePhys and the above networks.

Specifically, we first obtain the video feature extractor $E_{backbone}$ by removing the rPPG estimator (i.e., the last two linear layers of DeepPhys and EfficientPhys, the last two upsampling layers and the last convolutional layer of PhysFormer) of the backbone networks. Secondly, we replace the spatial-aware encoder (i.e., composed of video feature extractor $E_v$ and spatio-temporal encoder $E_{vf}$ together with the auxiliary prior branch) with the obtained video feature extractor $E_{backbone}$ to get the integrated network.

Table \ref{table:other_impl} presents the experimental results. For DeepPhys, the integrated network reduces MAE by 4.29 bpm and RMSE by 2.7 bpm. This paradigm also achieves significant improvement for EfficientPhys and PhysFormer. With improved performance and robustness, most end-to-end methods in the rPPG community could benefit from adopting this paradigm.

\subsection{Evaluation on Robustness}
\label{subsec:robust_study}

\begin{table}[t]
\centering
\caption{Study on the robustness against different types of visual interference for CodePhys and PhysFormer.}
\begin{tabular}{ccccccc}
\toprule
{Method} & {Visual Interference} & MAE$\downarrow$ & RMSE$\downarrow$\\

\midrule
\multirow{6}{*}{PhysFormer} & None & 4.82 & 8.17\\
& Motion Blur & 7.39 &  10.4\\
& Camera Noise & 7.58 & 10.3\\
& Varying Resolution & 6.31 & 9.51\\
& Occlusion & 8.56 & 10.7\\
& Varying Brightness & 7.47 & 10.4\\
\midrule
\multirow{6}{*}{CodePhys} & None & 4.30 & 7.29\\
& Motion Blur & 4.35 & 7.49\\
& Camera Noise & 4.57 & 7.38\\
& Varying Resolution & 4.62 & 7.78\\
& Occlusion & 5.19 & 8.13\\
& Varying Brightness & 5.33 & 8.64\\
\bottomrule
\end{tabular}
\label{table:robust_study}
\end{table}

\begin{figure}[t]
\centering
\includegraphics[width=1\columnwidth]{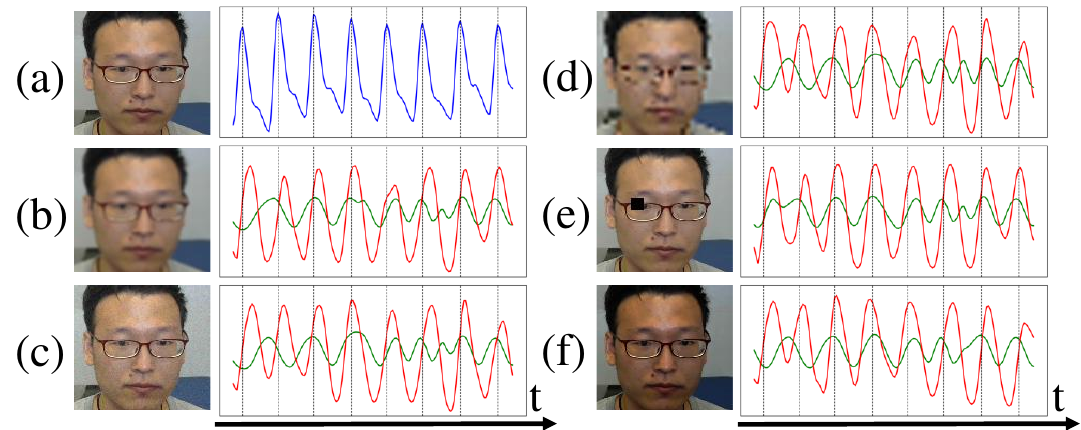}\vspace{-0.5em} 
\caption{Visualization of rPPG signals predicted by CodePhys (red) and PhysFormer (green) in case of video quality degradation, alongside with the GT-PPG (blue).}
\label{fig:video_degrade}
\end{figure}

Casting the rPPG estimation task to a code query task with the assistance of soft feature distillation guarantees the robust performance of CodePhys, even when videos are subject to visual interference. As shown in Fig. \ref{fig:video_degrade},  we simulate five possible categories of interference that may exist in real-world videos. The degradation of these interference changes randomly over time. Fig. \ref{fig:video_degrade}(a) denotes the original facial video and corresponding GT-PPG signal. Fig. \ref{fig:video_degrade}(b) uses Gaussian filtering to simulate defocus, where the kernel size of Gaussian blur is set to 5$\times$5, and sigma varies between 0.5 and 1.5. Fig. \ref{fig:video_degrade}(c) uses Gaussian noise to simulate camera noise, where the Gaussian noise is generated for each frame within the range of $[0,0.1]$. Fig. \ref{fig:video_degrade}(d) uses image interpolation to simulate varying resolutions, where the video resolutions vary between 32$\times$32 and 64$\times$64. Fig. \ref{fig:video_degrade}(e) uses facial area masks to simulate occlusion, where the facial mask is set to $\frac{H}{10}$$\times$$\frac{W}{10}$ and randomly positioned within the facial area. Fig. \ref{fig:video_degrade}(f) uses gamma correction to simulate varying brightness, where the gamma coefficient used in gamma correction varies within the range of $[0.5,1.5]$. We use CodePhys and the state-of-the-art method (i.e., PhysFormer) trained on the Fold-1 of VIPL-HR dataset to test their robustness. The quantitative results can be found in TABLE \ref{table:robust_study}. The severe degradation of video quality causes a noticeable decline in the performance of PhysFormer, while CodePhys still maintains accurate HR estimation. The above results demonstrate the remarkable robustness of our approach under various practical noise disturbances. The robustness mainly stems from the learning of noise-free latent PPG representation and the soft feature distillation loss.

\section{Conclusion}
\label{sec:conclusion}
In this paper, we propose a novel architecture CodePhys, which treats rPPG measurement as a code query task within a noise-free PPG feature space. With a tailor-made spatial-aware encoder network and soft feature distillation, CodePhys achieves significantly better results than existing state-of-the-art methods on benchmark datasets. However, research on resisting visual interference in rPPG measurement is still at an early stage. The datasets collected in a laboratory environment may not fully capture the variability of real-world applications. Additionally, the computational efficiency of CodePhys for real-time monitoring in practical settings requires further optimization. Therefore, future directions include: 1) Exploring additional cues under the noise-free PPG representation to further reduce the rPPG distortion; 2) Exploring the online learning or transfer learning approaches; and 3) Reducing the number of parameters in the modules of CodePhys to achieve faster inference speeds on edge devices.

\section*{References}

\bibliographystyle{IEEEtran}
\bibliography{reference}




\end{document}